%% file: main.tex
\documentclass[runningheads]{llncs}

\usepackage[mobile]{eccv}

\usepackage{eccvabbrv}

\usepackage{graphicx}
\usepackage{booktabs}

\usepackage[accsupp]{axessibility}  

\usepackage{hyperref}

\usepackage{graphicx}
\usepackage{xcolor}
\usepackage{amsmath}
\usepackage{amssymb}
\usepackage{booktabs}
\usepackage{times}
\usepackage{epsfig}
\usepackage[table,xcdraw]{xcolor}
\usepackage{multirow}
\usepackage{color}
\usepackage{arydshln}
\usepackage{float}
\usepackage{caption}
\usepackage{orcidlink}
\definecolor{red}{HTML}{FD654E}
\definecolor{purple}{HTML}{9295FF}
\definecolor{best}{HTML}{c1fed2}
\definecolor{green}{HTML}{20b74a}

\newif\ifarxiv
\arxivtrue      

\newcommand{\myref}[2]{%
\ifarxiv
    \ref{#1}%
\else
    #2%
\fi
}

\begin{document}

\title{Raw-JPEG Adapter: Efficient Raw Image Compression with JPEG}

\titlerunning{Raw-JPEG Adapter}

\author{Mahmoud Afifi\and
Ran Zhang\and
Michael S. Brown}

\authorrunning{M.~Afifi et al.}

\institute{AI Center-Toronto, Samsung Electronics\\
\vspace{1mm}
\email{m.3afifi@gmail.com}\\
\email{\{ran.zhang, michael.b1\}@samsung.com}}

\maketitle
\begin{center}
    \centering
    \captionsetup{type=figure} 
    \includegraphics[width=\textwidth]{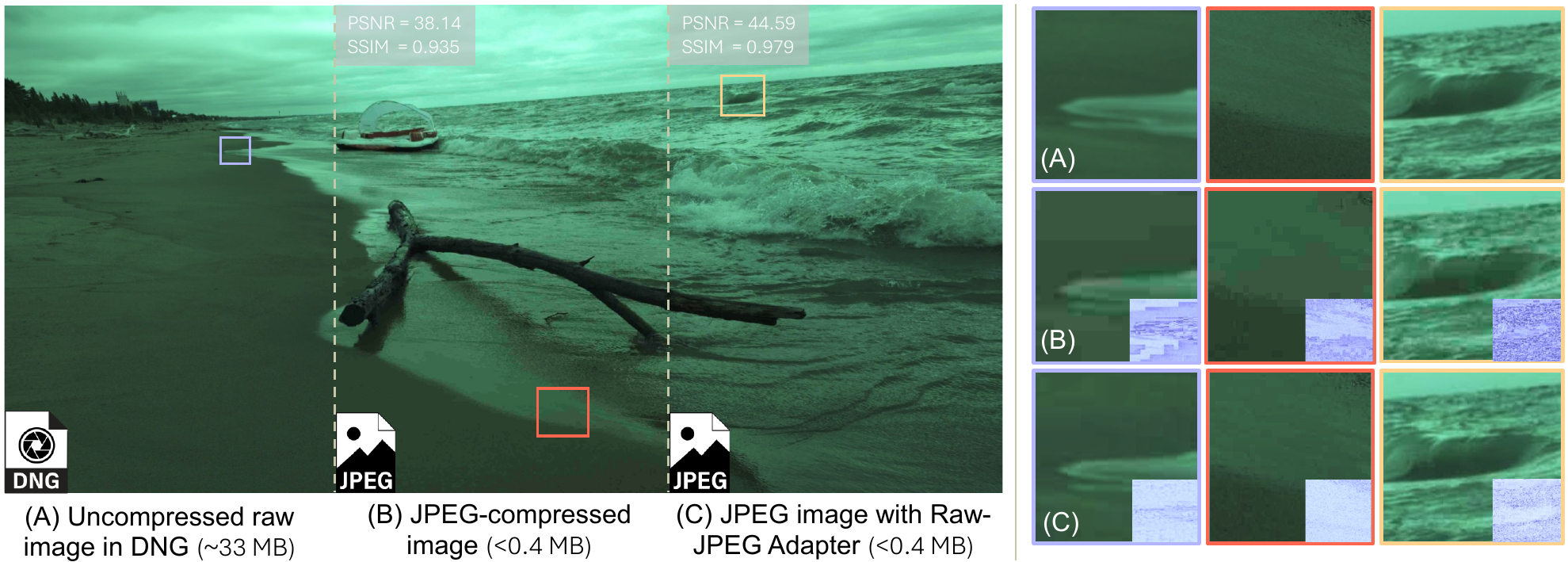}
    \vspace{-3mm}
\captionof{figure}{
We present Raw-JPEG Adapter, a lightweight, learnable pre-processing pipeline that adapts raw images before standard JPEG compression using spatial and optionally frequency domain transforms. The operations are fully invertible, with parameters fitting in the JPEG comment field ($<$64 KB), enabling accurate raw reconstruction after JPEG decoding and significantly reducing file size. In this figure, (A) shows the original raw (DNG), stored as JPEG with high compression (quality 25) without our method in (B), and with our method in (C). Error maps for (B) and (C) are shown on the right. All raw images shown in this paper are tone-mapped for visualization.\label{fig:teaser}}
    \vspace{-1mm}
\end{center}

\input{sec/0_abstract}

\input{sec/1_intro}
\input{sec/2_related_work}
\input{sec/3_method}

\input{sec/4_results}

\input{sec/5_conclusion}
\appendix
\input{sec/X_suppl}

\bibliographystyle{splncs04}
\bibliography{main}
\end{document}

%% file: sec/0_abstract.tex
\begin{abstract}
Digital cameras digitize scene light into linear raw representations, which the image signal processor (ISP) converts into display-ready outputs. While raw data preserves full sensor information---valuable for editing and vision tasks---formats such as Digital Negative (DNG) require large storage, making them impractical in constrained scenarios. In contrast, JPEG is a widely supported format, offering high compression efficiency and broad compatibility, but it is not well-suited for raw storage. This paper presents Raw-JPEG Adapter, a lightweight, learnable, and invertible pre-processing pipeline that adapts raw images for standard JPEG compression. Our method applies spatial and optional frequency-domain transforms, with compact parameters stored in the JPEG comment field, enabling accurate raw reconstruction. Experiments across multiple datasets show that our method achieves higher fidelity than direct JPEG storage, supports other codecs, and provides a favorable trade-off between compression ratio and reconstruction accuracy.
\end{abstract}

%% file: sec/1_intro.tex
\section{Introduction}
\label{sec:intro}


Modern digital cameras capture incoming light and digitize it into linear raw sensor measurements, which are subsequently processed by the camera’s image signal processor (ISP) to produce display-ready images (e.g., 8-bit sRGB). Raw data, however, is typically recorded at 12--14 bits per channel, preserving the full sensor information and offering greater flexibility for post-capture editing and computer vision applications \cite{schewe2012digital, cui2024raw, li2024efficient, berdan2025reraw, kee2025removing}.

Despite their benefits, raw images remain challenging to store in consumer workflows because of the absence of practical storage formats. The most widely used option is the Digital Negative (DNG) format \cite{adobe_dng_spec}, which encapsulates raw sensor data together with metadata. While DNG is lossless and widely supported in professional photography software, its file sizes are typically tens of megabytes per image, making it inefficient for large-scale storage or sharing. Alternative approaches include saving raw data in lossless 16-bit image formats such as PNG-16 or TIFF, which likewise preserve the original content but incur substantial storage overhead. Even with modern storage hardware, such file sizes are prohibitive for mobile capture, cloud synchronization, or large-scale datasets in computer vision research.

In principle, JPEG \cite{jpeg_standard}, the most widely used lossy image compression standard \cite{smith2025jpegs, hudson2017jpeg}, could be employed to reduce raw storage requirements. However, JPEG was fundamentally designed for 8-bit gamma-corrected images, whereas raw sensor data is linear and typically stored at higher bit-depths. Applying JPEG directly to raw signals results in severe degradation, including banding, clipping, and irreversible artifacts that preclude accurate raw reconstruction (see Fig.~\ref{fig:teaser}-B). By contrast, newer formats such as JPEG 2000, JPEG XL, and HEIC offer technical improvements but remain only sparsely adopted due to limited ecosystem and platform support \cite{sharinpix_heic_limits_2024, google_jpegxl_chrome_2022, sneyers_case_for_jpegxl_2022, lowe2009status}.

As a result, there remains no widely supported and efficient format for storing raw images at manageable file sizes without sacrificing fidelity. This motivates us to seek a solution that leverages the ubiquity of standard JPEG compression while addressing its shortcomings for raw storage. Such a solution should be accurate enough to overcome JPEG’s quantization and banding limitations, yet lightweight, requiring minimal processing time for both encoding and decoding. Importantly, it should maintain compatibility with the standard decoding pipeline, requiring only lightweight modifications. To this end, we propose Raw-JPEG Adapter, a learnable, compact, and fully invertible pre-processing pipeline that adapts raw images for JPEG storage. At decoding time, the pre-processing is inverted using only lightweight operations---without any reliance on deep models---to reconstruct the original raw data with high fidelity. Our method is both efficient and effective, delivering substantial improvements in compression efficiency and reconstruction accuracy compared to existing alternatives.

\noindent\\\textbf{Contributions:}
In this paper, we introduce a method for adapting raw images before JPEG storage, enabling high-fidelity reconstruction after decoding. Our method is a learnable and fully invertible pre-processing pipeline applied before standard JPEG compression. The pipeline learns the coefficients of invertible operators in a self-supervised manner, effectively mitigating JPEG artifacts and improving raw reconstruction accuracy. The method is lightweight ($\sim$37K parameters), incurs only $\sim$0.1s runtime overhead, and requires less than 64 KB to store the learned coefficients, which are embedded directly in the JPEG comment (COM) segment. We extensively validate our method across multiple datasets, demonstrating strong generalization and significant improvements over raw pre-processing baselines and state-of-the-art raw reconstruction methods. Finally, we show that the Raw-JPEG Adapter can also be applied with alternative compression schemes, yielding consistent improvements. 
\ifarxiv
\else
The code and trained models will be released upon acceptance.
\fi

%% file: sec/2_related_work.tex
\section{Related work}
\label{sec:related-work}
The focus of this paper \textit{is not} on developing a new raw compression algorithm, but rather on adapting existing compression schemes---particularly JPEG, the most widely adopted format in practice \cite{li2020survey}---to better preserve raw data fidelity. To situate our work, we briefly review related research in three areas: 1) raw data compression, 2) raw image reconstruction, and 3) bidirectional neural ISPs.

\subsection{Raw data compression}
\label{sec:related-work-raw-compression}

Prior work on raw image compression has largely focused on codec-specific engineering \cite{chung2022compression}. Such methods are commonly categorized as compression-first, which operate directly on mosaiced color filter array (CFA) data before coding, or demosaicing-first, which first convert to RGB and then apply handcrafted chroma subsampling or luma adjustment schemes \cite{modified_subsampling, CDM, OLM, chiu2014improved, chung2022compression}. Among demosaicing-first approaches, the Optimal Luma Modification (OLM) method adjusts the luma component so that, after upsampling and RGB conversion, the reconstructed image better matches the original \cite{OLM}. Another line of work adapts the JPEG XS codec \cite{descampe2021jpeg} for Bayer CFA data by replacing its component decorrelation stage with a nonlinear point transform and the Star-Tetrix transform, which approximates gamma-like behavior (2.2) while remaining hardware-friendly \cite{richter2021bayer}. 

Although effective, these approaches are mainly driven by real-time video transport and streaming use cases (e.g., broadcast, HDMI/SDI, VR/AR), where low latency and codec compatibility outweigh storage efficiency. In contrast, we focus on compact raw image storage for photography and computer vision, where archiving, portability, and ecosystem support are key. Our Raw-JPEG Adapter provides a lightweight, learnable, and invertible pre-processing pipeline that works seamlessly with standard JPEG, offering a storage-oriented alternative.



\subsection{Raw image reconstruction}
\label{sec:related-work-raw-reconstruction}

Raw image reconstruction methods aim to recover raw sensor data in post-capture from display-referred images (e.g., sRGB) \cite{punnappurath2019learning}. This line of work can be viewed as a form of compressed raw storage, where additional side information is stored to enable recovery from standard 8-bit images. The nature, scale, and role of this auxiliary information vary substantially across approaches. Early works operated under the 64\,KB JPEG comment constraint \cite{nguyen2016raw, nguyen2018raw}, using the side information to compensate for irreversible ISP operations applied before JPEG encoding. Subsequent methods stored sparse raw samples ($<$2\%) to assist reconstruction \cite{punnappurath2021spatially, nam2022learning, INF}. More recent learning-based approaches rely on richer latent features, often requiring 1--2\,MB of auxiliary data per image in addition to the sRGB image itself \cite{R2LCM1, R2LCM2}. In these settings, metadata plays a central role in reconstructing raw data from ISP-rendered images, as it is required to compensate for the irreversible transformations introduced by tone mapping, gamma correction, and other non-linear ISP operations.

Our method shares the high-level goal of enabling storage-efficient raw preservation, but differs fundamentally in formulation. Rather than reconstructing raw from ISP-processed sRGB with substantial auxiliary data, we bypass sRGB rendering and store the raw image directly in the JPEG format. The JPEG quality parameter provides a standard user-controlled size–quality trade-off. We additionally embed a small amount of side information (well below 64\,KB) in the COM segment; however, this information consists only of coefficients of the forward pre-processing operators that adapt raw data to JPEG quantization behavior. It is not used to compensate for irreversible ISP operations, nor does it encode latent representations of the image. Raw-JPEG therefore preserves raw content directly, while using lightweight operator coefficients solely to improve robustness to quantization, resulting in a simple and storage-efficient pipeline.

\subsection{Bidirectional neural ISPs}
\label{sec:related-work-raw-reconstruction}

Camera ISPs apply a sequence of highly nonlinear operators to raw images to generate display-referred outputs, typically 8-bit sRGB \cite{delbracio2021mobile}. Neural ISPs learn to approximate this pipeline using deep neural networks \cite{ignatov2022pynet, ignatov2022microisp, lite-isp, ren2025ispdiffuser}. Some work extends this idea by proposing bidirectional frameworks \cite{afifi2021cie, zamir2020cycleisp}, enabling an approximate inversion of the ISP to reconstruct raw data from sRGB without storing additional metadata. These methods either jointly train two networks to minimize both the display-image rendering loss and the raw reconstruction loss \cite{afifi2021cie}, or design invertible neural ISP blocks that explicitly allow reversing the rendering operations \cite{xing21invertible}. 

While these approaches provide a path to raw recovery from compact storage (i.e., saving only the display image in formats such as JPEG), the reconstructed raw is typically limited in accuracy due to the highly nonlinear operations applied during neural ISP rendering. In contrast, our method bypasses ISP rendering entirely: we store the raw data itself, transformed only by a lightweight pipeline before JPEG compression. This enables higher reconstruction fidelity while still keeping storage requirements affordable (see Fig.~\ref{fig:teaser}-C).

%% file: sec/3_method.tex
\section{Method}
\label{sec:method}

\begin{figure*}[!t]
\centering
\includegraphics[width=\linewidth]{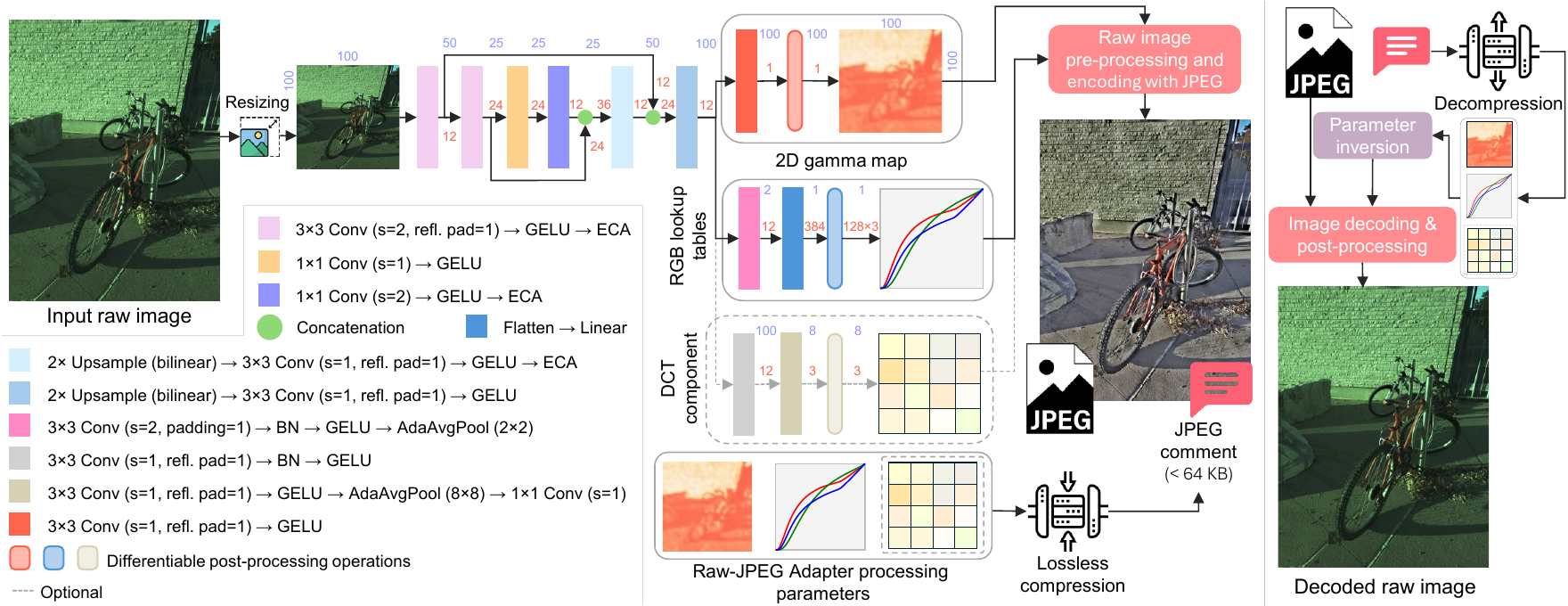}
\vspace{-4mm}
\caption{Our Raw-JPEG Adapter uses a lightweight network ($\sim$37K parameters) to process a thumbnail of the raw image and produce parameters for a pixel-wise gamma operator, RGB 1D tone-mapping lookup tables, and an optional DCT-based component applied globally in the frequency domain over 8$\times$8 blocks. These transformations are applied before saving the image as a JPEG, while the associated parameters are compressed and embedded in the JPEG file’s comment (COM) segment ($<$64 KB). All intermediate feature dimensionalities are annotated in the figure, with channel counts shown in {\textcolor{red}{red}} and spatial dimensions in {\textcolor{purple}{purple}}. At decoding time, the parameters are retrieved, inverted, and applied to the stored image to reconstruct the original raw content. During training, the JPEG step is replaced with a differentiable simulator, and the network is optimized in a self-supervised manner. \label{fig:main}}
\vspace{-2mm}
\end{figure*}

Given an input RGB demosaiced raw image $\mathbf{I} \in \mathbb{R}^{H \times W \times 3}$, after applying black-level normalization, our goal is to compress the image using a standard image codec (with a primary focus on JPEG), such that:

\begin{equation}
\label{eq:formulation}
\mathbf{I}' = \texttt{Dec}\left(\texttt{Enc}\left(\mathbf{I}\right)\right) \approx \mathbf{I},
\end{equation}

\noindent where $\texttt{Enc}(\cdot)$ and $\texttt{Dec}(\cdot)$ denote the encoder and decoder of the compression scheme, and $\mathbf{I}'$ is the reconstructed raw image. Since our design centers on JPEG, we first highlight the challenges of directly using JPEG for raw storage. Raw sensor data is typically recorded at 12–14 bits per channel, with most values concentrated in the lower intensity range due to the linear sensor response. The data also exhibits strong color casts arising from sensor sensitivity bias and scene illumination, both of which remain uncorrected at the raw stage. In contrast, JPEG is designed for 8-bit per channel RGB images in display-referred space (sRGB). Its pipeline applies color space conversion, chroma sub-sampling, blockwise $8\!\times\!8$ discrete cosine transform (DCT), and quantization using tables tuned for sRGB image statistics.

Directly saving raw data with JPEG therefore compresses 12–14 bits information to 8 bits, subjects the data to suboptimal bucket sizes misaligned with raw distributions. While the JPEG quality parameter $Q \in [1,100]$ controls the aggressiveness of quantization—higher $Q$ values reduce loss at the cost of larger file sizes—this trade-off is not well suited to raw images. Even at the highest setting ($Q=100$), banding in dark regions, clipping of highlights, and color distortions often remain because the quantization tables and color transforms are not adapted to sensor-domain data. Moreover, at lower $Q$ values, these artifacts become even more pronounced, severely degrading downstream rendering quality (see Fig.~\ref{fig:teaser}-B).

To improve the fidelity of the reconstructed raw image, we introduce invertible pre-processing operators applied to the raw image $\mathbf{I}$ before compression and inverted after decoding. Accordingly, Eq.~\ref{eq:formulation} is updated as:

\begin{equation}
\label{eq:ours}
\mathbf{\hat{I}} = F^{-1}\left(\texttt{Dec}\left(\texttt{Enc}\left(F\left(\mathbf{I}; \theta\right)\right)\right); \theta\right),
\end{equation}

\noindent where $F$ denotes the invertible pre-processing operators, parameterized by coefficients $\theta$, and $\mathbf{\hat{I}}$ is the reconstructed raw image by our method. 
These coefficients are compactly serialized using zlib compression with Base64 encoding and embedded in the JPEG comment segment, constrained to remain under 64 KB.

\subsection{Pre-encoding}
\label{sec:encoding}
Before JPEG compression, the raw image is processed by a pipeline of invertible pre-processing operators. The parameters of these operators are predicted by a learnable network (Sec.~\ref{sec:network}). The pipeline balances expressiveness with minimal storage overhead and negligible decoding cost. It consists of three components:

\vspace{-2mm}
\noindent\\\textbf{Channel-wise 1D look-up tables (LuTs):}  
1D LuTs are applied to each of the three color channels in the demosaiced raw image to modulate the global tone and intensity distribution. Each LuT is constrained to $128\!\times\!1$ values:
\begin{equation}
    \mathbf{I}^{\texttt{LuT}}_{c(x,y)} = \texttt{LuT}_c\!\left(\mathbf{I}_{c(x,y)}\right), \quad c \in \{R,G,B\}.
\end{equation}

\vspace{-2mm}
\noindent\\\textbf{Blockwise DCT component scaling:} 
The image is split into non-overlapping $8\!\times\!8$ blocks, and a single global $8\!\times\!8$ scaling matrix $\mathbf{S}$ is applied element-wise to the DCT coefficients of each block. The modified coefficients are then transformed back to the spatial domain via the inverse DCT:
\begin{equation}
    \mathbf{I}^{\texttt{DCT}}_b = \texttt{DCT}^{-1}\!\left(\mathbf{S} \odot \texttt{DCT}(\mathbf{I}^{\texttt{LuT}}_b)\right),
\end{equation}
where $b$ indexes the image blocks and $\odot$ denotes element-wise multiplication. Since training is performed on data from a specific camera, $\mathbf{S}$ implicitly adapts to that camera’s characteristics (e.g., sensor noise profile and color distribution). This frequency-domain adjustment is useful because JPEG applies fixed quantization tables that are not optimized for raw distributions. By selectively rescaling DCT coefficients before compression, $\mathbf{S}$ helps align the transformed raw data with JPEG's quantization behavior. A drawback, however, is that $\mathbf{S}$ encodes biases specific to the training camera, limiting generalization to new camera with different characteristics. For completeness, we therefore report results both with and without this component in our camera-specific and cross-camera evaluations (Sec.~\ref{sec:results}).

\vspace{-2mm}
\noindent\\\textbf{Pixel-wise gamma mapping:}  
A $100 \!\times\! 100$ gamma map $\boldsymbol{\Gamma}$ stores local gamma scalar factors, which are bilinearly upsampled to the input resolution and applied in a pixel-wise manner:
\begin{equation}
\label{eq:gamma}
    \mathbf{I}^{\Gamma}_{(x,y)} = {\mathbf{I}^{\texttt{DCT}}_{(x,y)}}^{\,\boldsymbol{\Gamma}_{(x,y)}}.
\end{equation}

Our pre-encoding pipeline is designed to overcome JPEG's bottlenecks. The 1D LuTs adjust tonal balance and redistribute intensity values, improving robustness to quantization from bit-depth reduction. The DCT scaling adapts frequency-domain statistics to camera characteristics, mitigating the mismatch with JPEG's quantization tables, which were tuned for natural images in display space. Finally, the pixel-wise gamma map applies local nonlinear corrections that help alleviate banding in dark regions, reduce color shifts, and further compensate for information loss due to reduced bit-depth. Together, these operators address the key limitations of applying JPEG to raw data, enabling high-fidelity reconstruction while remaining lightweight and fully invertible. Note that the DCT component is optional, as it naturally encodes camera-specific characteristics. If omitted, $\mathbf{I}^{\texttt{DCT}}$ is replaced with $\mathbf{I}^{\texttt{LuT}}$ in Eq.~\ref{eq:gamma}. 

\begin{figure*}[!t]
\centering
\includegraphics[width=\linewidth]{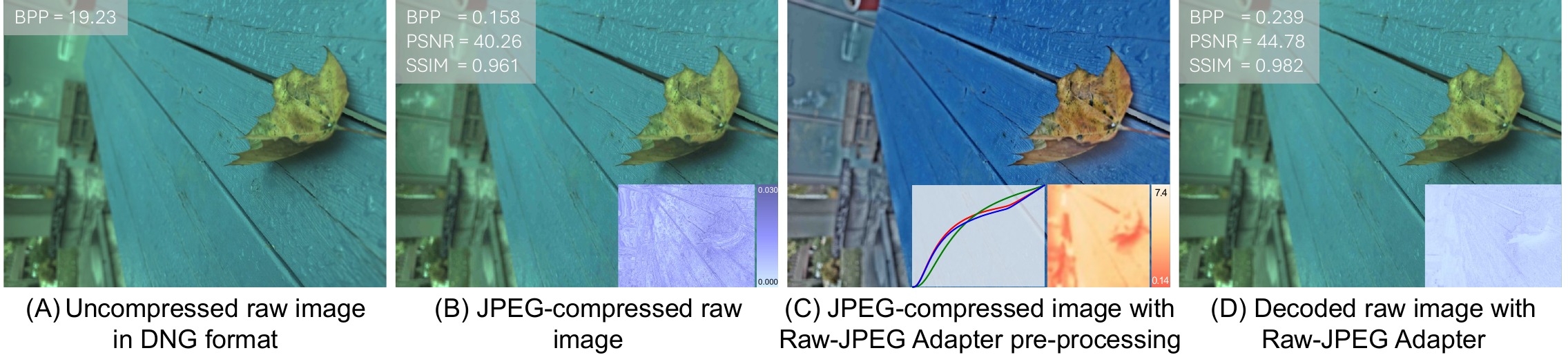}
\vspace{-4mm}
\caption{Qualitative example from the S24 test set \cite{S24} at JPEG quality 25. (A) Uncompressed demosaiced raw image. (B) JPEG-compressed raw without pre-/post-processing, with its error map. (C) JPEG image produced by our Raw-JPEG Adapter using the predicted RGB LuTs and gamma map. (D) Decoded raw image by our Raw-JPEG Adapter, along with the corresponding error map. \label{fig:qualtative_1}}
\vspace{-2mm}
\end{figure*}

\subsection{Post-decoding}
\label{sec:decoding}

At decoding time, we apply the inverse sequence of operations to reconstruct the raw image from the JPEG output. Given a decoded JPEG image $\mathbf{I}'$, the reconstruction proceeds in three steps:

\vspace{-2mm}
\noindent\\\textbf{Pixel-wise gamma inversion:}  
The gamma map $\boldsymbol{\Gamma}$ is bilinearly upsampled to the resolution of $\mathbf{I}'$, and an element-wise inverse gamma correction is applied:
\begin{equation}
    \mathbf{I}^{\Gamma}_{(x,y)} = \hat{\mathbf{I}}_{(x,y)}^{\,1/{\boldsymbol{\Gamma}}_{(x,y)}}.
\end{equation}

\vspace{-2mm}
\noindent\\\textbf{Inverse blockwise DCT scaling:}  
If the DCT scaling component was used during encoding, we invert it by dividing the blockwise DCT coefficients by the scaling matrix $\mathbf{S}$ before returning to the spatial domain:
\begin{equation}
    \mathbf{I}^{\texttt{IDCT}}_b = 
    \texttt{DCT}^{-1}\!\left(
        \texttt{DCT}(\mathbf{I}^{\Gamma}_b) \oslash \mathbf{S}
    \right),
\end{equation}
where $\oslash$ denotes element-wise division, $b$ indexes the $8 \!\times\! 8$ blocks, and $\mathbf{S}$ is the global scaling matrix estimated during encoding.  

\vspace{-2mm}
\noindent\\\textbf{Inverse channel-wise 1D LuTs:}  
Finally, channel-wise inverse LuTs are applied to restore the original distribution of each color channel:
\begin{equation}
    \mathbf{\hat{I}}_{c(x,y)} = \texttt{LuT}^{-1}_c\!\left(\mathbf{I}^{\texttt{IDCT}}_{c(x,y)}\right), \quad c \in \{R,G,B\}.
\end{equation}
If the DCT component is not used, $\mathbf{I}^{\texttt{IDCT}}$ is replaced by $\mathbf{I}^{\Gamma}$.  

Conceptually, the post-decoding pipeline starts from the $8$-bit quantized JPEG pixels and progressively restores the original raw domain. The inverse gamma step expands the limited $8$-bit range back toward the higher dynamic range of raw data, counteracting banding introduced during quantization. The DCT scaling then re-balances frequency-domain statistics that JPEG’s fixed quantization distorts, while the inverse LuTs recover the channel-wise tonal mapping applied at encoding. Together, these operations exactly invert the pre-processing steps. The parameters ($\boldsymbol{\Gamma}$, $\mathbf{S}$, \texttt{LuTs}) are compactly stored in the JPEG comment segment, and decoding requires only applying their inverse operators, without running any neural networks. As a result, the runtime overhead at decoding is negligible, while enabling high-fidelity reconstruction of the original raw image.

\subsection{Network Design}
\label{sec:network}

The parameters of our three operators ($\boldsymbol{\Gamma}$, $\mathbf{S}$, and \texttt{LuTs}) are predicted by a compact network, illustrated in Fig.~\ref{fig:main}. The architecture follows a convolutional encoder-decoder design with GELU activations~\cite{hendrycks2016gaussian}, ECA channel-attention blocks~\cite{wang2020eca}, and skip connections. The decoder branches into three lightweight heads, each predicting the parameters of one operator (two heads if the DCT component is omitted). The network operates on a thumbnail of the input image to produce parameters adapted to that image. To ensure stability and invertibility, the raw outputs are passed through differentiable post-processing that maps them into valid coefficient ranges.

\vspace{-2mm}
\noindent\\\textbf{Gamma map:} The predicted gamma values are stabilized via:
\begin{equation}
\boldsymbol{\Gamma}_{(x,y)} = \exp\left(2 \textit{} \tanh(g_{\theta_{(x,y)}})\right),
\end{equation}
where $g_\theta$ denotes the network output corresponding to the gamma parameters. This mapping constrains the predictions to the valid range $[0.14,7.4]$, ensuring strictly positive exponents while avoiding degenerate or excessively large corrections.

\vspace{-2mm}
\noindent\\\textbf{DCT scaling matrix:} Frequency-domain scaling is constrained as:
\begin{equation}
\mathbf{S} = \exp\left(0.7 \textit{} \tanh(s_\theta)\right),
\end{equation}
where $s_\theta$ denotes the network output corresponding to the DCT scaling matrix. This mapping constrains the coefficients to the range $[0.5, 2.0]$, enabling controlled frequency amplification or attenuation while avoiding instability.

\vspace{-2mm}
\noindent\\\textbf{Channel-wise 1D LuTs:}  
To guarantee monotonicity (and thus invertibility), each LuT is parameterized using unconstrained network outputs $h_\theta$, which are mapped to positive increments via \texttt{softplus}. The cumulative sum then enforces a strictly increasing mapping. Specifically, the sampled LuT values $\mathbf{L}$ are constructed as:
\begin{equation}
    \mathbf{L}(i) = \texttt{cumsum}\!\left(\texttt{softplus}(h_\theta)\right)_i,
\end{equation}
where $i$ indexes the discretized entries of the LuT. The resulting $\mathbf{L}$ represents the response values of the channel-wise look-up tables, which are then min–max normalized to $[0,1]$. This ensures that each channel-wise look-up table, $\texttt{LuT}_c$, represents a valid increasing mapping, while remaining differentiable and stable during training.

\subsection{Training}
\label{sec:training}

Our framework is trained in a self-supervised manner. Given an uncompressed raw training image $\mathbf{I}$, we predict the operator parameters $(\mathbf{S}, \mathbf{\Gamma}, \texttt{LuTs})$ using the network in Sec.~\ref{sec:network}. The image is transformed with these operators (Sec.~\ref{sec:encoding}), passed through a differentiable JPEG simulator~\cite{xing21invertible}, and inverted (Sec.~\ref{sec:decoding}) to yield the reconstruction $\hat{\mathbf{I}}$. The supervision signal is simply the original input $\mathbf{I}$, enabling end-to-end training without external ground truth pairs. 

To improve robustness across cameras with different sensor sensitivities and varying responses to incident light, we apply two lightweight augmentations during training (in addition to standard geometric augmentations):
1) intensity scaling, where the image is multiplied by a random factor in $[0.85,1.15]$ to simulate brightness variations, and  
2) color augmentation, where a random $3 \!\times\! 3$ matrix is applied to the RGB channels. To mimic the bias of real camera response functions toward the green channel---arising from sensor spectral sensitivities---we add a small fixed offset to the green row of the random color transformation matrix before normalization (i.e., +0.05). This enforces a slight green boost, reflecting the fact that most cameras capture proportionally more green light.

During training, we optimize the composite loss:
\begin{equation}
\label{eq:loss}
\mathcal{L} = \lambda_{\text{L1}} \, \mathcal{L}_{\text{L1}} +
              \lambda_{\text{SSIM}} \, \mathcal{L}_{\text{SSIM}} +
              \lambda_{\text{FFT}} \, \mathcal{L}_{\text{FFT}},
\end{equation}
where $\mathcal{L}_{\text{L1}}$ enforces pixel fidelity, $\mathcal{L}_{\text{SSIM}}$ promotes structural similarity, and $\mathcal{L}_{\text{FFT}}$ ensures frequency-domain consistency by applying an $\ell_1$ loss to the real and imaginary components of the 2D FFT of $\hat{\mathbf{I}}$ and $\mathbf{I}$.

%% file: sec/4_results.tex
\section{Experiments}
\label{sec:experiments}

\begin{figure*}[!t]
\centering
\includegraphics[width=\linewidth]{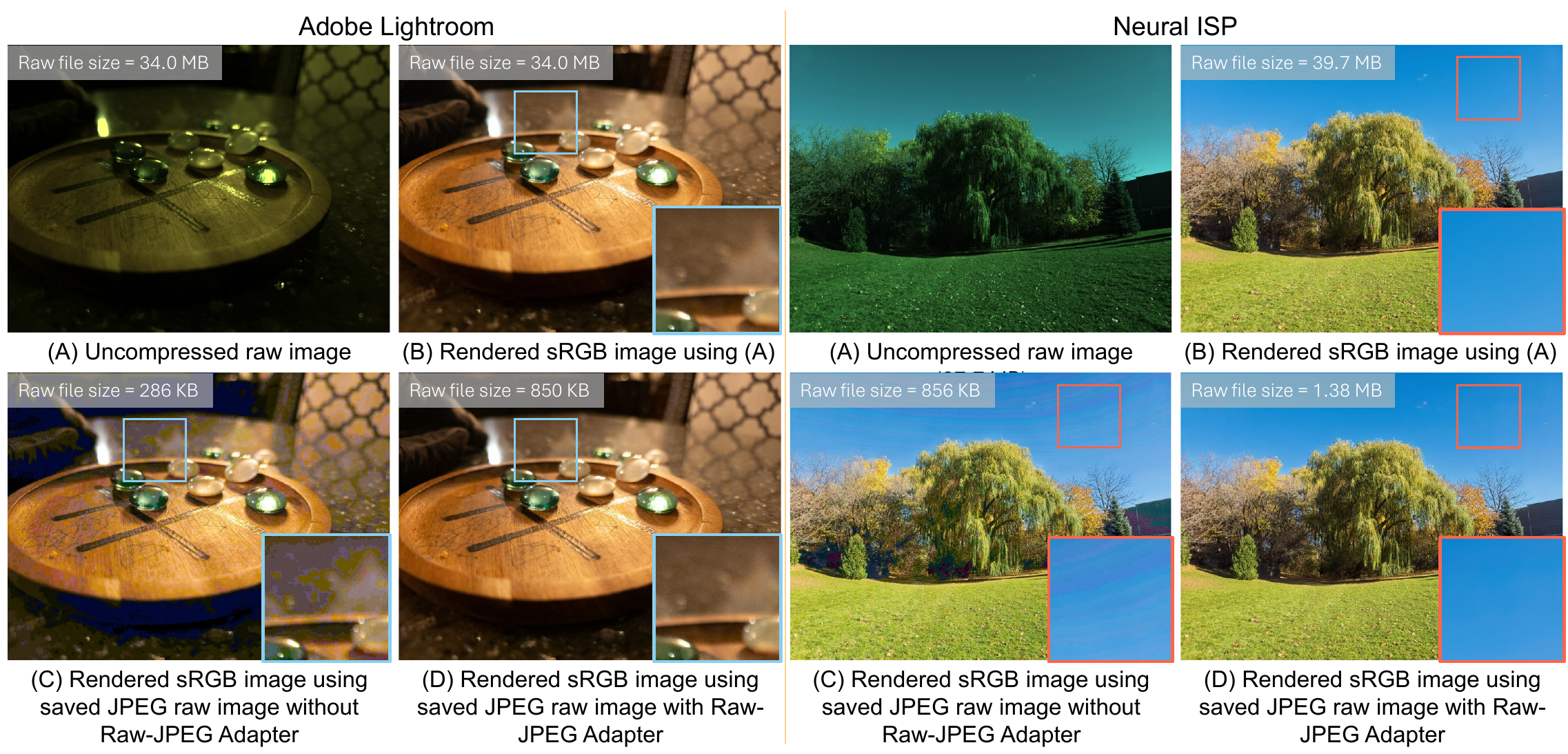}
\vspace{-4mm}
\caption{On the left and right, we show sRGB images rendered by Adobe Lightroom and LiteISP~\cite{lite-isp}, respectively, from: (B) the uncompressed raw image in (A) stored as a DNG file, (C) the JPEG-compressed raw image without our method, and (D) the JPEG-compressed raw image with our method. The raw JPEG images used to produce (C) and (D) were saved at quality level 50.
\label{fig:rendered_srgb}}
\vspace{-2mm}
\end{figure*}

We trained our model on the S24 training set~\cite{S24}, which contains 2,619 raw images from the Samsung S24 Ultra main camera. Training ran for 100 epochs with Adam~\cite{kingma2014adam} (learning rate $0.001$, $(\beta_1,\beta_2)=(0.9,0.999)$, weight decay $10^{-4}$). A cosine scheduler~\cite{loshchilov2016sgdr} decayed the rate to $10^{-5}$. Non-overlapping $512\times512$ patches were used. Loss weights were $\lambda_{\text{L1}}=1.0$, $\lambda_{\text{FFT}}=0.1$, and $\lambda_{\text{SSIM}}=0.1$.

As in-domain evaluation, we used the S24 test set (400 images). For cross-camera generalization, we tested on three external datasets: S7 (220 smartphone images) \cite{S7}, MIT-Adobe FiveK (5,000 images from 35 DSLR cameras) \cite{Adobe5K}, and NUS (1,736 images from 8 DSLR cameras) \cite{NUS}. All Bayer raws were demosaiced with Menon's algorithm \cite{menon2007demosaicing}.

We report results at JPEG qualities 100, 95, 75, 50, and 25, training a separate model for each. Comparisons include invertible pre-processing alternatives, metadata-based methods, and bidirectional neural ISPs. Our method \textit{is not} a new codec but a fidelity-oriented pre-processing stage for JPEG, which naturally inherits JPEG's storage efficiency. Bits-per-pixel (BPP) is determined by the quality setting, but our Raw-JPEG Adapter consistently achieves higher fidelity than alternatives while maintaining comparable BPP and significantly outperforming direct JPEG on raw.

For metrics, we compute PSNR, SSIM, and MS-SSIM~\cite{ssim, ms_ssim} against uncompressed raw, and measure compression ratio (CR) relative to 16-bit uncompressed raw PNGs. We report results with and without the DCT component and color/intensity augmentations to assess their impact (see Sec.~\myref{sec:ablations}{2.3} in the supp. materials for additional ablation studies).

\subsection{Results}
\label{sec:results}

\input{tables/results_0}
The results on the S24 test set \cite{S24} are shown in Table~\ref{tab:quality-100-supp} for JPEG quality 100. This setting represents the best-case scenario for storing raw images with JPEG, as the quantization tables are scaled such that no quantization loss occurs in the DCT coefficients. We compare our method against three pre-processing baselines: 
1) OLM~\cite{OLM}, applied prior to JPEG compression; 
2) raw$\leftrightarrow$sRGB mapping, where raw data are converted to sRGB using white-balance gains, a color correction matrix (CCM) provided in each DNG file, and a fixed 2.2 gamma correction; and 
3) fixed gamma correction, where a 2.2 gamma is applied before saving and inverted at decoding. 
For the raw$\leftrightarrow$sRGB baseline, the white-balance and CCM parameters are stored in the COM segment, similar to our method, and inverted at decoding time. For completeness, we also report results at additional JPEG quality levels in Table~\ref{tab:s24-jpeg-vs-ours}; see Fig.~\ref{fig:qualtative_1} for qualitative comparisons, with further results provided in the supp. material.

To ensure fair comparison under practical storage constraints, we further evaluate our method against standard JPEG compression under controlled bitrate settings. For each target BPP, we perform per-image rate control and evaluate at closely matched achieved bitrates. Under this setting, Raw-JPEG Adapter improves PSNR by approximately 2.6\,dB on average compared to direct JPEG compression at matched BPP levels. Additional details are provided in Sec.~\myref{supp-sec:fixed-bpp}{2.2} of the supp. material.

\input{tables/results_1}

\input{tables/deep_compression}

\input{tables/comparisons}

\input{tables/results_2}

We further examine our method in combination with deep image compression by replacing the JPEG codec with a learning-based alternative. Specifically, we adopt the LIC-TCM method~\cite{LIC}, a deep learning–based image compression model, in place of JPEG. We report results for our Raw-JPEG Adapter trained with a JPEG simulator, as well as after fine-tuning it with the pre-trained, fixed-weight LIC-TCM model. As shown in Table~\ref{tab:deep-compression}, our method achieves consistent improvements across different JPEG qualities and even when paired with deep image compression (see supp. material for a qualitative example). We also evaluated our model trained with the JPEG simulator on JPEG~2000, where it likewise showed improvements; additional details are provided in the supp. material.

When comparing our method with alternative approaches, including bidirectional neural ISPs~\cite{afifi2021cie, xing21invertible} that reconstruct raw from sRGB, and metadata-based methods~\cite{INF, R2LCM1, R2LCM2} that store auxiliary data alongside sRGB images, Table~\ref{tab:comparisons} shows that our method achieves the highest reconstruction accuracy while incurring minimal runtime overhead and maintaining comparable BPP.

In the results reported in Table~\ref{tab:comparisons}, the bidirectional neural ISPs~\cite{afifi2021cie, xing21invertible} were trained on the S24 dataset~\cite{S24}. Among the metadata-based methods, INF~\cite{INF} does not require training, as it performs self-supervised fitting of an implicit neural function at test time. The R2LCM method~\cite{R2LCM1, R2LCM2} was trained on the S24 dataset using paired JPEG-compressed sRGB images rendered with an advanced rendering pipeline that includes local tone mapping and complex color processing, along with the corresponding raw images. Both INF and R2LCM were evaluated on JPEG-compressed sRGB test images from the S24 dataset generated using the same rendering pipeline, reflecting practical imaging scenarios in which sRGB images are rendered through a complex rendering pipeline and subsequently JPEG-compressed. Additional discussion is provided in the supp.~material (Sec.~\myref{supp-sec:protocol}{1.4}). A radar chart comparison is shown in Fig.~\ref{fig:limitation_and_tradeoff}, and qualitative examples are included in the supp.~material.

We further evaluate our model trained on S24 images across different datasets to examine cross-camera generalization, as shown in Table~\ref{tab:cross-camera}. As described in Sec.~\ref{sec:method}, the DCT component is partially camera-specific, capturing sensor-dependent characteristics such as noise distribution. While this specialization can be advantageous for device-specific deployments, it generalizes less effectively across cameras. The results confirm this: the learned DCT component improves fidelity on the S24 dataset (the training camera; see Tables~\ref{tab:quality-100-supp} and \ref{tab:s24-jpeg-vs-ours}), but provides limited or no gains on other cameras, whereas the spatial-domain pipeline alone proves more robust. Qualitative examples are provided in the supp. material.

%% file: tables/results_0.tex
\begin{table}[t]
\centering
\caption{Quantitative results on the S24 test set \cite{S24}. We report average PSNR, SSIM, and MS-SSIM \cite{ssim, ms_ssim} between uncompressed raw images and decoded output at JPEG quality 100. Compression ratio (CR) is measured relative to the uncompressed 16-bit RGB PNG, and BPP corresponds to the JPEG file size. For our method, Raw-JPEG Adapter, we show results with and without the DCT component and our color/intensity augmentations. Best results are highlighted in \textcolor{green}{\textbf{green}}, and second-best results are shown in \textbf{bold}.\label{tab:quality-100-supp}}
\vspace{-1mm}
\scalebox{0.7}{
\begin{tabular}{|l|c|c|c|c|c|}
\hline
\multirow{2}{*}{\textbf{Method}} &
  \multicolumn{5}{c|}{\cellcolor[HTML]{fff0dd}\textbf{Quality = 100}} \\
\cline{2-6}
& \multicolumn{1}{c|}{\textbf{PSNR}} &
  \multicolumn{1}{c|}{\textbf{\begin{tabular}[c]{@{}c@{}}SSIM\\ ($\times$100)\end{tabular}}} &
  \multicolumn{1}{c|}{\textbf{\begin{tabular}[c]{@{}c@{}}MS-SSIM\\ ($\times$100)\end{tabular}}} &
  \multicolumn{1}{c|}{\textbf{BPP}} &
  \textbf{CR} \\
\hline
JPEG & \multicolumn{1}{c|}{46.00} & \multicolumn{1}{c|}{98.47} & \multicolumn{1}{c|}{99.72} & \multicolumn{1}{c|}{4.36} & 6.51 \\ \hline
JPEG + OLM \cite{OLM} & \multicolumn{1}{c|}{43.65} & \multicolumn{1}{c|}{97.74} & \multicolumn{1}{c|}{99.55} & \multicolumn{1}{c|}{4.21} & 6.73 \\ \hline
JPEG + raw$\leftrightarrow$sRGB & \multicolumn{1}{c|}{47.65} & \multicolumn{1}{c|}{99.25} & \multicolumn{1}{c|}{99.86} & \multicolumn{1}{c|}{6.09} & 4.70 \\ \hline
JPEG + fixed gamma & \multicolumn{1}{c|}{48.38} & \multicolumn{1}{c|}{99.27} & \multicolumn{1}{c|}{\textbf{99.87}} & \multicolumn{1}{c|}{5.25} & 5.47 \\ 
\hdashline
Raw-JPEG Adapter (w/o DCT, w/o aug) & \multicolumn{1}{c|}{48.49} & \multicolumn{1}{c|}{\textbf{99.28}} & \multicolumn{1}{c|}{\textbf{99.87}} & \multicolumn{1}{c|}{5.28} & 5.43 \\ \hline
Raw-JPEG Adapter (w/o DCT, w/ $\text{  }$ aug) & \multicolumn{1}{c|}{48.44} & \multicolumn{1}{c|}{\textbf{99.28}} & \multicolumn{1}{c|}{\textbf{99.87}} & \multicolumn{1}{c|}{5.30} & 5.41 \\ \hline

Raw-JPEG Adapter (w/ $\text{  }$ DCT, w/o aug) & 
\multicolumn{1}{c|}{\textbf{48.99}} & \multicolumn{1}{c|}{\cellcolor{best}\textbf{99.35}} & \multicolumn{1}{c|}{\cellcolor{best}\textbf{99.89}} & \multicolumn{1}{c|}{5.28} & 5.40 \\ \hline

Raw-JPEG Adapter (w/ $\text{  }$ DCT, w/ $\text{  }$ aug) & \multicolumn{1}{c|}{\cellcolor{best}\textbf{49.04}} & \multicolumn{1}{c|}{\cellcolor{best}\textbf{99.35}} & \multicolumn{1}{c|}{\cellcolor{best}\textbf{99.89}} & \multicolumn{1}{c|}{5.28} & 5.41 \\
\hline
\end{tabular}
}
\end{table}

%% file: tables/results_1.tex
\begin{table*}[t]
\centering
\caption{Quantitative results on the S24 test set \cite{S24} across different JPEG quality levels. For Raw-JPEG Adapter, we report results with and without DCT and color/intensity augmentations. Best results are highlighted in \textcolor{green}{\textbf{green}}, and second-best in \textbf{bold}.\label{tab:s24-jpeg-vs-ours}}\vspace{-1mm}

\scalebox{0.49}{
\begin{tabular}{|l|ccccc|ccccc|ccccc|ccccc|}
\hline
\multirow{2}{*}{\textbf{Method}} &
  \multicolumn{5}{c|}{\cellcolor[HTML]{fed18d}\textbf{Quality = 25}} &
  \multicolumn{5}{c|}{\cellcolor[HTML]{ff9484}\textbf{Quality = 50}} &
  \multicolumn{5}{c|}{\cellcolor[HTML]{d5effe}\textbf{Quality = 75}} &
  \multicolumn{5}{c|}{\cellcolor[HTML]{CBCEFB}\textbf{Quality = 95}} \\
\cline{2-21}
& \multicolumn{1}{c|}{\textbf{PSNR}} &
  \multicolumn{1}{c|}{\textbf{\begin{tabular}[c]{@{}c@{}}SSIM\\ ($\times$100)\end{tabular}}} &
  \multicolumn{1}{c|}{\textbf{\begin{tabular}[c]{@{}c@{}}MS-SSIM\\ ($\times$100)\end{tabular}}} &
  \multicolumn{1}{c|}{\textbf{BPP}} &
  \textbf{CR} &
  \multicolumn{1}{c|}{\textbf{PSNR}} &
  \multicolumn{1}{c|}{\textbf{\begin{tabular}[c]{@{}c@{}}SSIM\\ ($\times$100)\end{tabular}}} &
  \multicolumn{1}{c|}{\textbf{\begin{tabular}[c]{@{}c@{}}MS-SSIM\\ ($\times$100)\end{tabular}}} &
  \multicolumn{1}{c|}{\textbf{BPP}} &
  \textbf{CR} &
  \multicolumn{1}{c|}{\textbf{PSNR}} &
  \multicolumn{1}{c|}{\textbf{\begin{tabular}[c]{@{}c@{}}SSIM\\ ($\times$100)\end{tabular}}} &
  \multicolumn{1}{c|}{\textbf{\begin{tabular}[c]{@{}c@{}}MS-SSIM\\ ($\times$100)\end{tabular}}} &
  \multicolumn{1}{c|}{\textbf{BPP}} &
  \textbf{CR} &
  \multicolumn{1}{c|}{\textbf{PSNR}} &
  \multicolumn{1}{c|}{\textbf{\begin{tabular}[c]{@{}c@{}}SSIM\\ ($\times$100)\end{tabular}}} &
  \multicolumn{1}{c|}{\textbf{\begin{tabular}[c]{@{}c@{}}MS-SSIM\\ ($\times$100)\end{tabular}}} &
  \multicolumn{1}{c|}{\textbf{BPP}} &
  \textbf{CR} \\
\hline
JPEG &
  \multicolumn{1}{c|} {36.58} & \multicolumn{1}{c|}{90.08} & \multicolumn{1}{c|}{95.79} & \multicolumn{1}{c|}{0.25} & 121.15 &
  \multicolumn{1}{c|}{39.02} & \multicolumn{1}{c|}{93.57} & \multicolumn{1}{c|}{97.74} & \multicolumn{1}{c|}{0.36} & 86.57 &
  
\multicolumn{1}{c|}{40.75} & \multicolumn{1}{c|}{95.40} & \multicolumn{1}{c|}{98.62} & \multicolumn{1}{c|}{0.56} & 56.42 &
  
  \multicolumn{1}{c|}{43.60} & \multicolumn{1}{c|}{97.21} & \multicolumn{1}{c|}{99.35} & \multicolumn{1}{c|}{1.67} & 17.78 \\
\hline

JPEG + OLM \cite{OLM} &
  \multicolumn{1}{c|}{36.53} & \multicolumn{1}{c|}{90.09} & \multicolumn{1}{c|}{95.77} & \multicolumn{1}{c|}{0.25} & 121.87 &
  \multicolumn{1}{c|}{38.90} & \multicolumn{1}{c|}{93.53} & \multicolumn{1}{c|}{97.71} & \multicolumn{1}{c|}{0.36} & 87.43 &

\multicolumn{1}{c|}{40.33} & \multicolumn{1}{c|}{95.17} & \multicolumn{1}{c|}{98.56} & \multicolumn{1}{c|}{0.55} & 57.20 &

  \multicolumn{1}{c|}{42.42} & \multicolumn{1}{c|}{96.84} & \multicolumn{1}{c|}{99.25} & \multicolumn{1}{c|}{1.64} & 17.98 \\ \hline

JPEG + raw$\leftrightarrow$sRGB &
  \multicolumn{1}{c|}{39.41} & \multicolumn{1}{c|}{95.25} & \multicolumn{1}{c|}{98.39} & \multicolumn{1}{c|}{0.33} & 92.99 &
  \multicolumn{1}{c|}{41.26} & \multicolumn{1}{c|}{96.58} & \multicolumn{1}{c|}{99.07} & \multicolumn{1}{c|}{0.53} & 60.23 &

\multicolumn{1}{c|}{42.63} & \multicolumn{1}{c|}{97.37} & \multicolumn{1}{c|}{99.39} & \multicolumn{1}{c|}{0.85} & 37.15 &

  \multicolumn{1}{c|}{45.42} & \multicolumn{1}{c|}{98.52} & \multicolumn{1}{c|}{99.62} & \multicolumn{1}{c|}{2.58} & 11.83 \\ \hline

JPEG + fixed gamma &
  \multicolumn{1}{c|}{38.30} & \multicolumn{1}{c|}{94.51} & \multicolumn{1}{c|}{97.82} & \multicolumn{1}{c|}{0.30} & 101.32 &
  \multicolumn{1}{c|}{40.74} & \multicolumn{1}{c|}{96.20} & \multicolumn{1}{c|}{98.81} & \multicolumn{1}{c|}{0.47} & 68.00 &
\multicolumn{1}{c|}{42.43} & \multicolumn{1}{c|}{97.16} & \multicolumn{1}{c|}{99.25} & \multicolumn{1}{c|}{0.73} & 43.06 &

  \multicolumn{1}{c|}{45.42} & \multicolumn{1}{c|}{98.42} & \multicolumn{1}{c|}{99.65} & \multicolumn{1}{c|}{2.15} & 14.06 \\ \hdashline

Raw-JPEG Adapter (w/o DCT, w/o aug) &
  \multicolumn{1}{c|}{39.35} & \multicolumn{1}{c|}{95.71} & \multicolumn{1}{c|}{98.69} & \multicolumn{1}{c|}{0.48} & 63.78 &
\multicolumn{1}{c|}{41.69} & \multicolumn{1}{c|}{96.91} & \multicolumn{1}{c|}{99.22} & \multicolumn{1}{c|}{0.70} & 44.11 &

\multicolumn{1}{c|}{43.31} & \multicolumn{1}{c|}{97.61} & \multicolumn{1}{c|}{99.46} & \multicolumn{1}{c|}{1.03} & 30.19 &

\multicolumn{1}{c|}{45.97} & \multicolumn{1}{c|}{98.58} & \multicolumn{1}{c|}{99.71} & \multicolumn{1}{c|}{2.40} & 12.58  \\ \hline

Raw-JPEG Adapter (w/o DCT, w/ $\text{  }$ aug) &
\multicolumn{1}{c|}{39.49} & \multicolumn{1}{c|}{95.70} & \multicolumn{1}{c|}{98.69} & \multicolumn{1}{c|}{0.47} & 65.13 &
\multicolumn{1}{c|}{41.73} & \multicolumn{1}{c|}{96.87} & \multicolumn{1}{c|}{99.20} & \multicolumn{1}{c|}{0.69} & 45.10 &

\multicolumn{1}{c|}{43.36} & \multicolumn{1}{c|}{97.60} & \multicolumn{1}{c|}{99.46} & \multicolumn{1}{c|}{1.01} & 30.75 &

\multicolumn{1}{c|}{46.12} & \multicolumn{1}{c|}{\cellcolor{best}\textbf{98.61}} & \multicolumn{1}{c|}{\cellcolor{best}\textbf{99.72}} & \multicolumn{1}{c|}{2.46} & 12.29 \\ \hline

Raw-JPEG Adapter (w/ $\text{ }$ DCT, w/o aug) &
\multicolumn{1}{c|}{\textbf{39.70}} & \multicolumn{1}{c|}{\textbf{95.83}} & \multicolumn{1}{c|}{\textbf{98.77}} & \multicolumn{1}{c|}{0.49} & 62.92 &
\multicolumn{1}{c|}{\cellcolor{best}\textbf{41.92}} & \multicolumn{1}{c|}{\cellcolor{best}\textbf{96.99}} & \multicolumn{1}{c|}{\cellcolor{best}\textbf{99.26}} & \multicolumn{1}{c|}{0.71} & 43.43 &

\multicolumn{1}{c|}{\textbf{43.54}} & \multicolumn{1}{c|}{\cellcolor{best}\textbf{97.68}} & \multicolumn{1}{c|}{\cellcolor{best}\textbf{99.49}} & \multicolumn{1}{c|}{1.04} & 29.82 &
\multicolumn{1}{c|}{\cellcolor{best}\textbf{46.22}} & \multicolumn{1}{c|}{\textbf{98.60}} & \multicolumn{1}{c|}{\cellcolor{best}\textbf{99.72}} & \multicolumn{1}{c|}{2.35} & 12.87  \\ \hline

Raw-JPEG Adapter (w/ $\text{ }$ DCT, w/ $\text{  }$ aug) &
\multicolumn{1}{c|}{\cellcolor{best}\textbf{39.82}} & \multicolumn{1}{c|}{\cellcolor{best}\textbf{95.88}} & \multicolumn{1}{c|}{\cellcolor{best}\textbf{98.78}} & \multicolumn{1}{c|}{0.48} & 63.84 &
\multicolumn{1}{c|}{\textbf{41.88}} & \multicolumn{1}{c|}{\textbf{96.98}} & \multicolumn{1}{c|}{\textbf{99.25}} & \multicolumn{1}{c|}{0.71} & 43.73 &

\multicolumn{1}{c|}{\cellcolor{best}\textbf{43.58}} & \multicolumn{1}{c|}{\textbf{97.67}} & \multicolumn{1}{c|}{\textbf{99.48}} & \multicolumn{1}{c|}{1.04} & 29.85 &

\multicolumn{1}{c|}{\textbf{46.19}} & \multicolumn{1}{c|}{98.58} & \multicolumn{1}{c|}{\textbf{99.71}} & \multicolumn{1}{c|}{2.32} & 13.02 \\
\hline
\end{tabular}
}\vspace{-1mm}
\end{table*}

%% file: tables/deep_compression.tex
\begin{table}[t]
\centering
\caption{Quantitative results on the S24 test set \cite{S24} using the deep learning-based LIC-TCM method \cite{LIC} for compression instead of JPEG. We evaluate our method with/without tuning for LIC-TCM. 
Best results are highlighted in \textcolor{green}{\textbf{green}}.\label{tab:deep-compression}}

\vspace{-2mm}
\scalebox{0.7}{
\begin{tabular}{|l|c|c|c|c|c|}
\hline
\multicolumn{1}{|c|}{\textbf{Pre-/post-processing}}
& \multicolumn{1}{c|}{\textbf{PSNR}} &
  \multicolumn{1}{c|}{\textbf{\begin{tabular}[c]{@{}c@{}}SSIM\\ ($\times$100)\end{tabular}}} &
  \multicolumn{1}{c|}{\textbf{\begin{tabular}[c]{@{}c@{}}MS-SSIM\\ ($\times$100)\end{tabular}}} &
  \multicolumn{1}{c|}{\textbf{BPP}} &
  \textbf{CR} \\
\hline
None & \multicolumn{1}{c|}{42.95} & \multicolumn{1}{c|}{96.56} & \multicolumn{1}{c|}{99.03} & \multicolumn{1}{c|}{0.31} & 160.47
\\ \hline

Fixed gamma & \multicolumn{1}{c|}{43.55} & \multicolumn{1}{c|}{97.38} & \multicolumn{1}{c|}{99.32} & \multicolumn{1}{c|}{0.63} & 90.27 \\ \hdashline

Ours (w/o DCT) & \multicolumn{1}{c|}{45.11} & \multicolumn{1}{c|}{98.12} & \multicolumn{1}{c|}{99.58} & \multicolumn{1}{c|}{1.16} & 42.14 \\ \hline

Ours (w/$\text{ }$ DCT) & \multicolumn{1}{c|}{45.11} & \multicolumn{1}{c|}{98.12} & \multicolumn{1}{c|}{99.58} & \multicolumn{1}{c|}{1.16} & 42.14 \\ \hline

Ours (w/o DCT) - tuned & \multicolumn{1}{c|}{45.43} & \multicolumn{1}{c|}{98.53} & \multicolumn{1}{c|}{99.61} & \multicolumn{1}{c|}{1.17} & 41.52 \\ \hline

Ours (w/$\text{ }$ DCT) - tuned & \multicolumn{1}{c|}{\cellcolor{best}\textbf{46.03}} & \multicolumn{1}{c|}{\cellcolor{best}\textbf{98.75}} & \multicolumn{1}{c|}{\cellcolor{best}\textbf{99.82}} & \multicolumn{1}{c|}{1.26} & 40.97 \\ \hline

\end{tabular}
}
\vspace{-3mm}

\end{table}

%% file: tables/comparisons.tex
\begin{table}[t]
\centering
\caption{Comparisons with alternative methods on the S24 test set \cite{S24}. 
We compare Raw-JPEG Adapter against bidirectional neural ISPs \cite{afifi2021cie, xing21invertible}, metadata-based methods \cite{INF, R2LCM1, R2LCM2}, and our method applied with JPEG or deep compression \cite{LIC}. 
We report PSNR, SSIM, MS-SSIM \cite{ssim, ms_ssim}, BPP (total storage, with/without sRGB when applicable), and raw reconstruction runtime (seconds). 
Best results are in \textcolor{green}{\textbf{green}}, second-best in \textbf{bold}.\label{tab:comparisons}}\vspace{-1mm}

\scalebox{0.7}{
\begin{tabular}{|l|c|c|c|c|c|}
\hline
\textbf{Method} &
  \textbf{PSNR} &
  \textbf{\begin{tabular}[c]{@{}c@{}}SSIM\\ ($\times$100)\end{tabular}} &
  \textbf{\begin{tabular}[c]{@{}c@{}}MS-SSIM\\ ($\times$100)\end{tabular}} &
  \textbf{BPP} &
  \textbf{\begin{tabular}[c]{@{}c@{}}Runtime\\ overhead\end{tabular}} \\ \hline
CIE XYZ Net \cite{afifi2021cie}  &  20.47    & 81.61 & 91.84 & (2.340 / 0.000) &  0.301 \\ \hline
Invertible ISP \cite{xing21invertible}     & 43.71 & 98.40 & 99.56 & (2.340 / 0.000) &  7.932 \\ \hline
INF \cite{INF}                 & 41.24 & 96.29 & 98.56 & (3.764 /  0.776) & 25.93 \\ \hline
R2LCM (drop 0) \cite{R2LCM1, R2LCM2}     & 46.01 & 98.49 & 99.67 & (3.929 / 1.701) & 2.815 \\ \hline
R2LCM (drop 1)  \cite{R2LCM1, R2LCM2}    & 44.8 &  98.34 & 99.61 & (3.867 / 1.639) & 2.786 \\ \hline
R2LCM (drop 2) \cite{R2LCM1, R2LCM2}     & 38.42 & 96.28 & 98.55 & (2.831 / 0.602) &  2.261 \\ \hline
R2LCM (drop 4) \cite{R2LCM1, R2LCM2}     & 33.65 & 94.56 & 96.28 & (2.379 / 0.150)  & 2.023 \\ \hline
JPEG + ours (quality = 50) & 41.92 & 96.99 & 99.26 & (0.713 / 0.713) & 0.120 \\ \hline
JPEG + ours (quality = 75) & 43.58 & 97.67 & 99.48 & (1.036 / 1.036) & 0.120 \\ \hline
JPEG + ours (quality = 95) & \cellcolor{best}\textbf{46.22} & \textbf{98.60} & \textbf{99.72}  & (2.345 / 2.345) & 0.120 \\ \hline
LIC-TCM \cite{LIC} + ours & \textbf{46.03} & \cellcolor{best}\textbf{98.75} & \cellcolor{best}\textbf{99.82} &  (1.260 / 1.260) &  0.120 \\ \hline
\end{tabular}}
\vspace{-2mm}
\end{table}

%% file: tables/results_2.tex
\begin{table}[t]
\centering
\caption{Quantitative results on cross-camera generalization to unseen cameras during training. We report results across different JPEG quality levels. Best results are highlighted in \textcolor{green}{\textbf{green}}.\label{tab:cross-camera}}\vspace{-2mm}
\scalebox{0.7}{
\begin{tabular}{|c|ccc|ccc|ccc|ccc|}
\hline
\multicolumn{13}{|c|}{\cellcolor[HTML]{e6e5e1}\textbf{S7 dataset \cite{S7} (220 images taken by 1 smartphone camera)}} \\
\hline
\multirow{2}{*}{\textbf{Method}} &
  \multicolumn{3}{c|}{\cellcolor[HTML]{fed18d}\textbf{Quality = 25}} &
  \multicolumn{3}{c|}{\cellcolor[HTML]{ff9484}\textbf{Quality = 50}} &
  \multicolumn{3}{c|}{\cellcolor[HTML]{d5effe}\textbf{Quality = 75}} &
  \multicolumn{3}{c|}{\cellcolor[HTML]{CBCEFB}\textbf{Quality = 95}} \\
\cline{2-13}

\multicolumn{1}{|c|}{\textbf{}} &
  \multicolumn{1}{c|}{\textbf{PSNR}} &
  \multicolumn{1}{c|}{\textbf{\begin{tabular}[c]{@{}c@{}}SSIM \\ ($\times$100)\end{tabular}}} &
  \multicolumn{1}{c|}{\textbf{BPP}} &
  \multicolumn{1}{c|}{\textbf{PSNR}} &
  \multicolumn{1}{c|}{\textbf{\begin{tabular}[c]{@{}c@{}}SSIM \\ ($\times$100)\end{tabular}}} &
  \multicolumn{1}{c|}{\textbf{BPP}} &
  \multicolumn{1}{c|}{\textbf{PSNR}} &
  \multicolumn{1}{c|}{\textbf{\begin{tabular}[c]{@{}c@{}}SSIM \\ ($\times$100)\end{tabular}}} &
  \multicolumn{1}{c|}{\textbf{BPP}} &
    \multicolumn{1}{c|}{\textbf{PSNR}} &
  \multicolumn{1}{c|}{\textbf{\begin{tabular}[c]{@{}c@{}}SSIM \\ ($\times$100)\end{tabular}}} &
  \multicolumn{1}{c|}{\textbf{BPP}}
  \\ \hline

\multicolumn{1}{|c|}{JPEG} &
\multicolumn{1}{c|}{37.54} & \multicolumn{1}{c|}{85.81} & \multicolumn{1}{c|}{0.18} & \multicolumn{1}{c|}{39.58} & \multicolumn{1}{c|}{90.02} & \multicolumn{1}{c|}{0.27} & \multicolumn{1}{c|}{40.94} & \multicolumn{1}{c|}{92.55} & \multicolumn{1}{c|}{0.47} & \multicolumn{1}{c|}{43.84} & \multicolumn{1}{c|}{96.41} & 1.75 
  \\ \hline

  \multicolumn{1}{|c|}{Raw$\leftrightarrow$sRGB} &

\multicolumn{1}{c|}{38.29} & \multicolumn{1}{c|}{91.75} & \multicolumn{1}{c|}{0.33} & \multicolumn{1}{c|}{39.13} & \multicolumn{1}{c|}{93.45} & \multicolumn{1}{c|}{0.64} & \multicolumn{1}{c|}{39.95} & \multicolumn{1}{c|}{95.05} & \multicolumn{1}{c|}{1.16} & \multicolumn{1}{c|}{42.76} & \multicolumn{1}{c|}{98.00} & 3.70  
  \\ \hline

  \multicolumn{1}{|c|}{Fixed gamma} &
\multicolumn{1}{c|}{39.77} & \multicolumn{1}{c|}{91.20} & \multicolumn{1}{c|}{0.29} & \multicolumn{1}{c|}{41.26} & \multicolumn{1}{c|}{93.13} & \multicolumn{1}{c|}{0.52} & \multicolumn{1}{c|}{42.46} & \multicolumn{1}{c|}{94.85} & \multicolumn{1}{c|}{0.93} & \multicolumn{1}{c|}{46.38} & \multicolumn{1}{c|}{98.08} & 2.98 
  \\ \hdashline

  \multicolumn{1}{|c|}{Ours (w/ $\text{  }$ DCT)} &
\multicolumn{1}{c|}{39.86} & \multicolumn{1}{c|}{\cellcolor{best}\textbf{93.05}} & \multicolumn{1}{c|}{0.51} & \multicolumn{1}{c|}{40.85} & \multicolumn{1}{c|}{\cellcolor{best}\textbf{94.62}} & \multicolumn{1}{c|}{0.86} & \multicolumn{1}{c|}{42.15} & \multicolumn{1}{c|}{\cellcolor{best}\textbf{96.05}} & \multicolumn{1}{c|}{1.38} & \multicolumn{1}{c|}{44.99} & \multicolumn{1}{c|}{98.21} & 3.18
  \\ \hline

  \multicolumn{1}{|c|}{Ours (w/o DCT)} &
\multicolumn{1}{c|}{\cellcolor{best}\textbf{41.04}} & \multicolumn{1}{c|}{92.98} & \multicolumn{1}{c|}{0.49} & \multicolumn{1}{c|}{\cellcolor{best}\textbf{42.23}} & \multicolumn{1}{c|}{94.56} & \multicolumn{1}{c|}{0.81} & \multicolumn{1}{c|}{\cellcolor{best}\textbf{43.46}} & \multicolumn{1}{c|}{95.98} & \multicolumn{1}{c|}{1.34} & \multicolumn{1}{c|}{\cellcolor{best}\textbf{47.24}} & \multicolumn{1}{c|}{\cellcolor{best}\textbf{98.39}} & 3.37 

  \\ \hline

\hline
\multicolumn{13}{|c|}{\cellcolor[HTML]{e6e5e1}\textbf{MIT-Adobe 5K dataset \cite{Adobe5K} (5,000 images taken by 35 DSLR cameras)}} \\
\hline

\multicolumn{1}{|c|}{JPEG} &
\multicolumn{1}{c|}{38.96} & \multicolumn{1}{c|}{91.13} & \multicolumn{1}{c|}{0.19} & \multicolumn{1}{c|}{41.90} & \multicolumn{1}{c|}{95.02} & \multicolumn{1}{c|}{0.25} & \multicolumn{1}{c|}{43.91} & \multicolumn{1}{c|}{96.69} & \multicolumn{1}{c|}{0.35} & \multicolumn{1}{c|}{46.38} & \multicolumn{1}{c|}{97.83} & 1.14 

  \\ \hline

  \multicolumn{1}{|c|}{Raw$\leftrightarrow$sRGB} &
\multicolumn{1}{c|}{38.42} & \multicolumn{1}{c|}{96.26} & \multicolumn{1}{c|}{0.25} & \multicolumn{1}{c|}{39.69} & \multicolumn{1}{c|}{97.04} & \multicolumn{1}{c|}{0.38} & \multicolumn{1}{c|}{40.43} & \multicolumn{1}{c|}{97.46} & \multicolumn{1}{c|}{0.63} & \multicolumn{1}{c|}{41.73} & \multicolumn{1}{c|}{98.17} & 2.18 
  \\ \hline

  \multicolumn{1}{|c|}{Fixed gamma} &
\multicolumn{1}{c|}{41.88} & \multicolumn{1}{c|}{96.66} & \multicolumn{1}{c|}{0.23} & \multicolumn{1}{c|}{44.42} & \multicolumn{1}{c|}{97.72} & \multicolumn{1}{c|}{0.33} & \multicolumn{1}{c|}{46.09} & \multicolumn{1}{c|}{98.27} & \multicolumn{1}{c|}{0.53} & \multicolumn{1}{c|}{48.96} & \multicolumn{1}{c|}{99.05} & 1.80 
  \\ \hdashline

  \multicolumn{1}{|c|}{Ours (w/ $\text{  }$ DCT)} &
\multicolumn{1}{c|}{40.96} & \multicolumn{1}{c|}{\cellcolor{best}\textbf{97.47}} & \multicolumn{1}{c|}{0.36} & \multicolumn{1}{c|}{42.47} & \multicolumn{1}{c|}{\cellcolor{best}\textbf{98.11}} & \multicolumn{1}{c|}{0.52} & \multicolumn{1}{c|}{43.99} & \multicolumn{1}{c|}{98.53} & \multicolumn{1}{c|}{0.78} & \multicolumn{1}{c|}{45.73} & \multicolumn{1}{c|}{99.11} & 1.95 
  \\ \hline

  \multicolumn{1}{|c|}{Ours (w/o DCT)} &
\multicolumn{1}{c|}{\cellcolor{best}\textbf{43.07}} & \multicolumn{1}{c|}{97.42} & \multicolumn{1}{c|}{0.35} & \multicolumn{1}{c|}{\cellcolor{best}\textbf{45.21}} & \multicolumn{1}{c|}{98.10} & \multicolumn{1}{c|}{0.50} & \multicolumn{1}{c|}{\cellcolor{best}\textbf{46.75}} & \multicolumn{1}{c|}{\cellcolor{best}\textbf{98.54}} & \multicolumn{1}{c|}{0.76} & \multicolumn{1}{c|}{\cellcolor{best}\textbf{49.67}} & \multicolumn{1}{c|}{\cellcolor{best}\textbf{99.20}} & 2.08 
  \\ \hline

\hline
\multicolumn{13}{|c|}{\cellcolor[HTML]{e6e5e1}\textbf{NUS dataset \cite{NUS} (1,736 images taken by 8 DSLR cameras)}} \\
\hline

\multicolumn{1}{|c|}{JPEG} &
\multicolumn{1}{c|}{37.92} & \multicolumn{1}{c|}{89.56} & \multicolumn{1}{c|}{0.21} & \multicolumn{1}{c|}{40.54} & \multicolumn{1}{c|}{93.78} & \multicolumn{1}{c|}{0.29} & \multicolumn{1}{c|}{42.32} & \multicolumn{1}{c|}{95.83} & \multicolumn{1}{c|}{0.42} & \multicolumn{1}{c|}{44.59} & \multicolumn{1}{c|}{97.15} & 1.22 
  \\ \hline

  \multicolumn{1}{|c|}{Raw$\leftrightarrow$sRGB} &
\multicolumn{1}{c|}{40.10} & \multicolumn{1}{c|}{96.97} & \multicolumn{1}{c|}{0.30} & \multicolumn{1}{c|}{41.77} & \multicolumn{1}{c|}{97.87} & \multicolumn{1}{c|}{0.44} & \multicolumn{1}{c|}{42.86} & \multicolumn{1}{c|}{98.34} & \multicolumn{1}{c|}{0.66} & \multicolumn{1}{c|}{44.49} & \multicolumn{1}{c|}{98.88} & 1.92 
  \\ \hline

  \multicolumn{1}{|c|}{Fixed gamma} &
\multicolumn{1}{c|}{41.26} & \multicolumn{1}{c|}{96.46} & \multicolumn{1}{c|}{0.28} & \multicolumn{1}{c|}{43.58} & \multicolumn{1}{c|}{97.61} & \multicolumn{1}{c|}{0.41} & \multicolumn{1}{c|}{45.16} & \multicolumn{1}{c|}{98.19} & \multicolumn{1}{c|}{0.62} & \multicolumn{1}{c|}{47.68} & \multicolumn{1}{c|}{98.84} & 1.80 
  \\ \hdashline

  \multicolumn{1}{|c|}{Ours (w/ $\text{  }$ DCT)} &

\multicolumn{1}{c|}{41.48} & \multicolumn{1}{c|}{\cellcolor{best}\textbf{97.50}} & \multicolumn{1}{c|}{0.50} & \multicolumn{1}{c|}{43.17} & \multicolumn{1}{c|}{\cellcolor{best}\textbf{98.14}} & \multicolumn{1}{c|}{0.68} & \multicolumn{1}{c|}{44.64} & \multicolumn{1}{c|}{\cellcolor{best}\textbf{98.51}} & \multicolumn{1}{c|}{0.93} & \multicolumn{1}{c|}{46.25} & \multicolumn{1}{c|}{98.92} & 1.97 
  \\ \hline

\multicolumn{1}{|c|}{Ours (w/o DCT)} &
\multicolumn{1}{c|}{\cellcolor{best}\textbf{42.45}} & \multicolumn{1}{c|}{97.41} & \multicolumn{1}{c|}{0.49} & \multicolumn{1}{c|}{\cellcolor{best}\textbf{44.45}} & \multicolumn{1}{c|}{98.06} & \multicolumn{1}{c|}{0.66} & \multicolumn{1}{c|}{\cellcolor{best}\textbf{45.97}} & \multicolumn{1}{c|}{98.49} & \multicolumn{1}{c|}{0.91} & \multicolumn{1}{c|}{\cellcolor{best}\textbf{48.32}} & \multicolumn{1}{c|}{\cellcolor{best}\textbf{98.97}} & 2.11 
  \\ \hline
\end{tabular}
}
\vspace{-2mm}
\end{table}

%% file: sec/5_conclusion.tex
\section{Conclusion and discussion}
\label{sec:conclusion}

\begin{figure}[t]
\centering
\includegraphics[width=\linewidth]{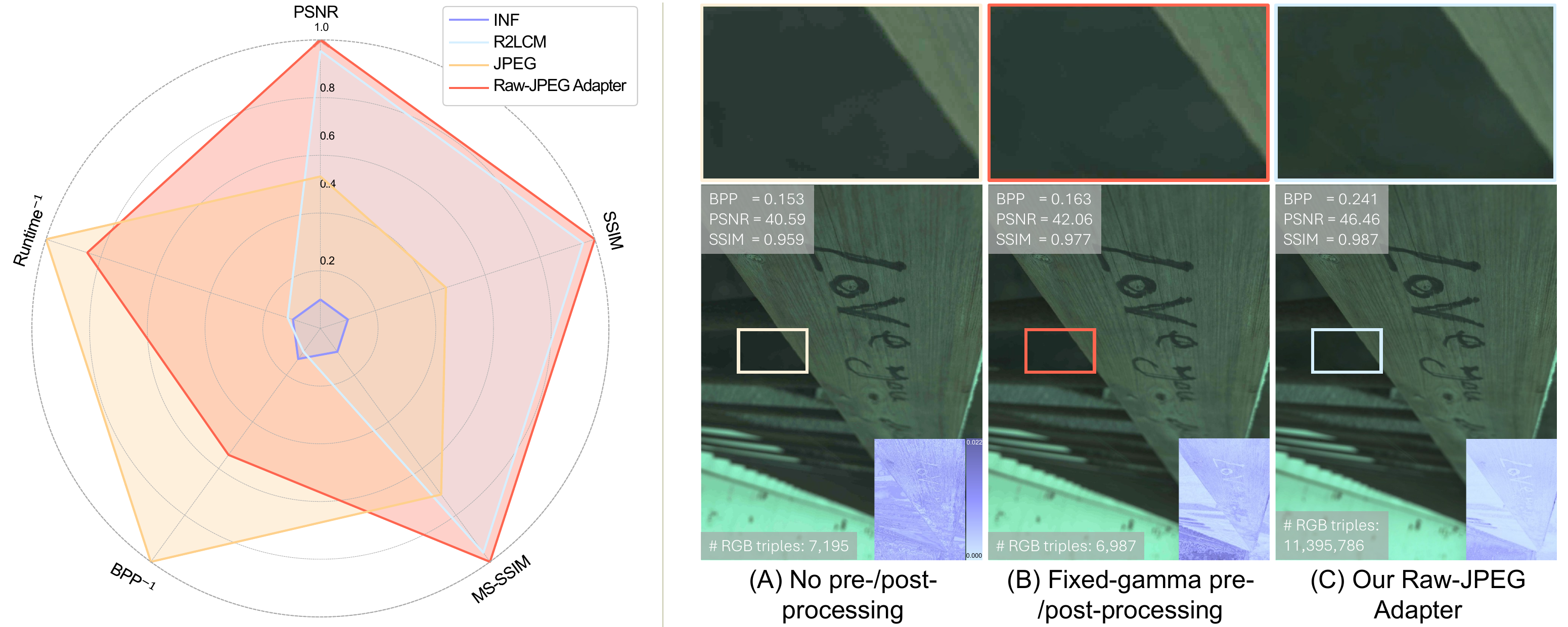}
\vspace{-3mm}
\caption{Raw-JPEG Adapter improves raw image fidelity under JPEG compression while maintaining a favorable trade-off between compression efficiency, reconstruction quality, and runtime. 
Right: Visual comparison showing that, although our method significantly reduces artifacts, minor compression artifacts may still persist. 
(A) JPEG-decoded raw image without pre-/post-processing and its error map. 
(B) JPEG-decoded raw image with fixed 2.2 gamma applied before encoding and inverted after decoding, with its error map. 
(C) Decoded raw image produced by our method, along with the corresponding error map. 
Left: Comparison of compression methods in terms of normalized metrics (PSNR, SSIM, MS-SSIM, inverted BPP, and inverted runtime overhead) on the S24 dataset~\cite{S24}, including JPEG, INF~\cite{INF}, R2LCM~\cite{R2LCM1,R2LCM2}, and  Raw-JPEG Adapter.}
\label{fig:limitation_and_tradeoff}
\vspace{-1mm}
\end{figure}

We presented a lightweight method that significantly improves raw image storage using JPEG, leveraging its lossy compression for small file sizes while enhancing fidelity and reconstruction accuracy. The results in Sec.~\ref{sec:results} demonstrate consistent improvements over alternative approaches and strong generalization across datasets. Furthermore, we showed that our framework integrates seamlessly with alternative compression methods, while remaining lightweight with negligible runtime overhead.

\noindent\\\textbf{Applications:}
Our method enables storing raw images at much smaller sizes while still supporting post-capture re-rendering and editing. To demonstrate this, we evaluated a pair of images from the S24 test set~\cite{S24} compressed at JPEG quality 50, both with and without our method, and compared the rendered sRGB outputs against those obtained from uncompressed raw data. For rendering, we used LiteISP~\cite{lite-isp}, trained on S24 raw images, and also tested Adobe Lightroom’s raw-to-sRGB pipeline. Since Lightroom requires proprietary decoding, we adopted the approach of~\cite{seo2023graphics2raw}: our decoded JPEGs (with and without our method) were stored as temporary DNGs by replacing the image data, which were then rendered to sRGB in Lightroom. As shown in Fig.~\ref{fig:rendered_srgb}, our method produces perceptually similar results to uncompressed DNGs ($>$30 MB) while reducing storage to under 2~MB. In contrast, directly storing raw data as JPEG introduces visible artifacts. Additional examples and discussion are provided in the supp. material.

\noindent\\\textbf{Limitations and trade-offs:}
Despite its improvements, our method does not fully eliminate compression artifacts. 
Compared to direct JPEG decoding (Fig.~\ref{fig:limitation_and_tradeoff}-A) and the fixed 2.2 gamma baseline (Fig.~\ref{fig:limitation_and_tradeoff}-B), artifacts are clearly reduced, yet subtle distortions may still persist (Fig.~\ref{fig:limitation_and_tradeoff}-C). Our method introduces a small BPP overhead, since storing pre-processing parameters in the COM segment (typically $\sim$40\,KB, always $<$64\,KB) slightly increases file size. In practice, this overhead is negligible relative to uncompressed DNG files (tens of MBs) and is justified by the fidelity gains. Controlled bitrate experiments (Sec.~\myref{supp-sec:fixed-bpp}{2.2} of the supp. material) further confirm that our method consistently outperforms direct JPEG compression at matched BPP levels. Furthermore, conventional BPP does not fully reflect perceptual quality: our method reconstructs a substantially wider range of unique RGB triples (i.e., richer tonal distributions) than baseline JPEGs, as shown in Fig.~\ref{fig:limitation_and_tradeoff}. 
To better capture this effect, Sec.~\myref{supp-sec:weighted-bpp}{2.1} of the supp. material reports a weighted BPP metric that accounts for tonal diversity, providing a fairer measure of storage cost versus reconstruction quality. More broadly, the results in Fig.~\ref{fig:limitation_and_tradeoff} highlight the central trade-off addressed in this work: enabling efficient raw storage at manageable file sizes while preserving high-fidelity reconstruction and maintaining compatibility with standard JPEG decoding. 
Our Raw-JPEG Adapter achieves a favorable balance among compression efficiency, reconstruction quality, and runtime overhead compared to existing alternatives, while requiring only lightweight processing during both encoding and decoding.

%% file: sec/X_suppl.tex
\ifarxiv
\clearpage
\begin{center}
    {\Large\bfseries Supplementary Material}\\[1em]
\end{center}
\vspace{1em}
\fi

\section{Additional details}
\label{supp-sec:results}

\subsection{Blockwise DCT transformation} 
As described in the main paper, we optionally learn a global discrete cosine transform (DCT) component, represented as a set of $8\!\times\!8$ scale coefficients. Given an input raw image $\mathbf{I} \in \mathbb{R}^{H\!\times\!W\!\times\!3}$, after applying the 1D LuTs, we obtain $\mathbf{I}^\texttt{LuT}$. We first pad $\mathbf{I}^\texttt{LuT}$ so that both height and width are divisible by the block size $B=8$. The image is then partitioned into non-overlapping $B\!\times\!B$ patches, and we apply the 2D DCT to each patch independently. The 1D DCT basis of size $B$ is defined as:
\begin{equation}
\label{eq:dct_alpha}
\alpha(k) =
\begin{cases}
B^{-1/2}, & k=0,\\
(2/B)^{1/2}, & k=1,\dots,B-1,
\end{cases}
\end{equation}
\begin{equation}
\label{eq:dct_basis}
\begin{aligned}
\mathbf{D}_{k,i} &= \alpha(k)\cos\!\left(\tfrac{\pi}{2B}(2i+1)k\right), \\
&\quad i=0,\dots,B-1.
\end{aligned}
\end{equation}
where $\mathbf{D} \in \mathbb{R}^{B\!\times\!B}$. For a block $\mathbf{X} \in \mathbb{R}^{B\!\times\!B}$, the 2D DCT coefficients are obtained as:
\begin{equation}
\label{eq:dct2}
\mathbf{Y} = \mathbf{D} \, \mathbf{X} \, \mathbf{D}^\top,
\end{equation}
where $\mathbf{Y} \in \mathbb{R}^{B \times B}$ contains the DCT coefficients. 
When learning the optional global DCT component, we scale $\mathbf{Y}$ element-wise by a learnable tensor $\mathbf{S} \in \mathbb{R}^{B\!\times\!B}$ to obtain $\mathbf{\hat{Y}}$. The inverse transform is performed using the 2D inverse DCT:
\begin{equation}
\label{eq:idct2}
\mathbf{\hat{X}} = \mathbf{D}^\top \, \mathbf{\hat{Y}} \, \mathbf{D}.
\end{equation}

Finally, all reconstructed patches $\mathbf{\hat{X}}$ are reassembled via folding to recover the full image $\mathbf{I}^\texttt{DCT} \in \mathbb{R}^{H\!\times\!W\!\times\!3}$.

\subsection{ECA block}
In our network design, we adopt the Efficient Channel Attention (ECA) block~\cite{wang2020eca} as a lightweight channel-attention module. Our implementation follows the standard design of global average pooling, followed by a 1D convolution and sigmoid gating, and uses a fixed odd kernel size (rather than the adaptive kernel selection in~\cite{wang2020eca}); no dimensionality reduction or spatial attention is used.

\subsection{Differentiable JPEG simulator}
\label{sec:diff-jpeg}
During training, our method requires a fully differentiable JPEG simulation that mimics the degradation introduced by JPEG encoding. In the literature, a few differentiable JPEG simulators have been proposed (e.g., \cite{salamah2025jpeg, xing21invertible}). In our experiments, we adopt the differentiable JPEG simulator of \cite{xing21invertible}, which implements the full JPEG pipeline in a differentiable manner. An input RGB image is first transformed to YCbCr, followed by $4{:}2{:}0$ chroma subsampling, blockwise $8\!\times\!8$ DCT, and quantization of the DCT coefficients using luminance/chroma tables $T$ scaled by a quality-dependent factor $f(Q)$:
\begin{equation}
f(Q) = 
\begin{cases}
\dfrac{50}{Q}, & Q < 50, \\[8pt]
\dfrac{200 - 2Q}{100}, & Q \geq 50,
\end{cases}
\end{equation}
where $Q \in [1,100]$ is the JPEG quality parameter. Quantization of the DCT coefficient $C_{u,v}$ with table entry $T_{u,v}$ is then performed as
\begin{equation}
\hat{C}_{u,v} = \mathrm{round}\!\left(\frac{C_{u,v}}{T_{u,v}\, f(Q)}\right).
\end{equation}

To enable backpropagation, the rounding operator can be replaced with a differentiable approximation based on a truncated Fourier series:
\begin{equation}
\mathrm{round}(z) \approx z - \frac{1}{\pi}\sum_{n=1}^{N}\frac{(-1)^{n+1}}{n}\sin(2\pi n z),
\end{equation}
where $N$ controls the approximation accuracy. Decompression applies the inverse operations: de-quantization, inverse DCT, block merging, chroma upsampling, and YCbCr $\to$ RGB conversion. For ablation studies (Sec.~\ref{sec:ablations}), we replace this module with a simpler uniform quantization layer that directly simulates 8-bit rounding in RGB space:
\begin{equation}
\hat{x} = \frac{\mathrm{round}(255 \textit{ } x)}{255}, \quad x \in [0,1].
\end{equation}
In this simplified case, the color transform, chroma sub-sampling, block partitioning, and DCT transformations are omitted, leaving only the effect of scalar quantization.

\subsection{Evaluation protocol}
\label{supp-sec:protocol}

\begin{figure}[!h]
\centering
\includegraphics[width=\linewidth]{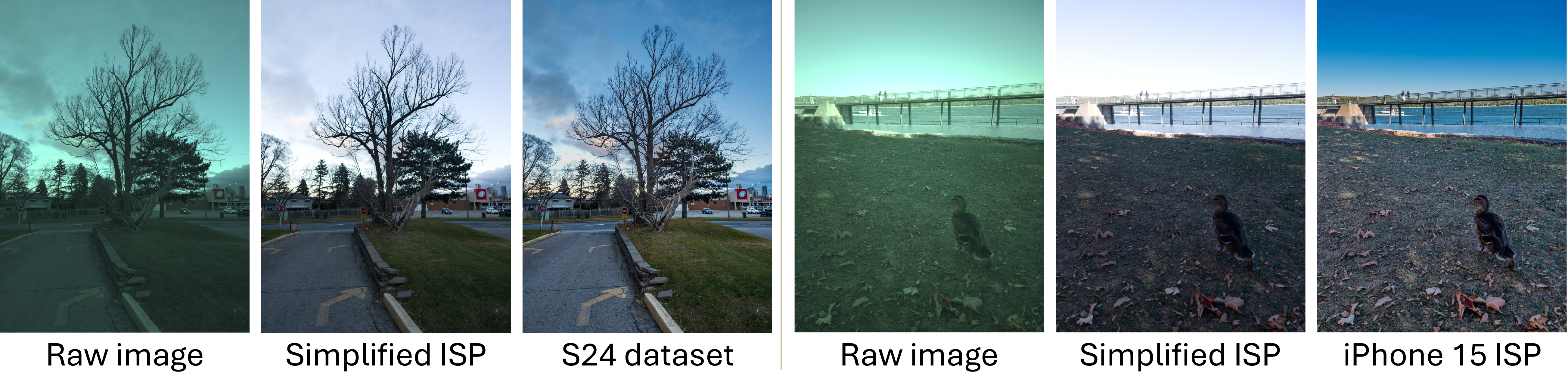}
\vspace{-4mm}
\caption{Comparison of raw image renderings produced by a simplified software ISP~\cite{karaimer2016software}, the S24 dataset rendering pipeline~\cite{S24}, and a modern commercial camera ISP (iPhone 15). The S24 rendering exhibits more complex tone mapping and color processing compared to the simplified ISP, resulting in a setting that more closely reflects practical camera pipelines.}
\label{fig:simple_rendering}
\vspace{-1mm}
\end{figure}

We clarify the differences between our evaluation protocol and those adopted in prior raw reconstruction works such as INF~\cite{INF} and R2LCM~\cite{R2LCM1, R2LCM2}. First, in prior sRGB-to-raw works (e.g.,~\cite{R2LCM1, R2LCM2}), the input sRGB images are uncompressed (PNG or TIFF). For reference, a 12\,MP sRGB PNG image typically occupies $\sim$20\,MB, while a TIFF file may reach 50--60\,MB per sRGB image. These sRGB images are rendered using a simplified software ISP composed primarily of global operators (following~\cite{punnappurath2021spatially, nam2022learning}) and stored without lossy compression, which results in very high PSNR values.

In contrast, our evaluation protocol in the main paper (Table~~\myref{tab:comparisons}{4}) reconstructs raw images from JPEG-compressed sRGB inputs. This setting more closely reflects practical camera pipelines, where sRGB outputs are typically stored in compressed form. Importantly, R2LCM itself reports a substantial performance drop when compressed sRGB inputs are used: in Table~3 of~\cite{R2LCM1}, PSNR decreases to 41--43.5\,dB at JPEG quality~90, consistent with the range reported in our paper.

Second, prior evaluations rely on a simplified rendering pipeline based on global transformations only \cite{karaimer2016software}. In contrast, the S24 dataset~\cite{S24} used in our experiments employs a high-quality rendering pipeline that includes local tone mapping and more complex color processing (see Fig.~\ref{fig:simple_rendering}). This increases the intrinsic ambiguity of sRGB-to-raw inversion and makes reconstruction substantially more challenging, while better reflecting realistic deployment scenarios.

\section{Additional results}
\label{supp-sec:results}

\subsection{Weighted bits per pixel (wBPP)}
\label{supp-sec:weighted-bpp}

\input{tables/wBPP}

In the main paper, we report the bits per pixel (BPP) of the compressed files. As shown in the results, our method introduces an increase in BPP due to the pre-processing and operator coefficients stored in the JPEG comment segment. When combined with more advanced compression algorithms, such as the deep learning LIC-TCM model~\cite{LIC}, our method achieves lower BPP while maintaining higher reconstruction accuracy. 

However, the conventional definition of BPP---the total number of bits used to store a compressed image divided by the number of pixels---does not account for tonal diversity in the reconstructed image. Two images with identical BPP can differ substantially in perceptual quality. In particular, an image that preserves richer tonal variations (i.e., more unique RGB triplets) is likely to provide higher raw reconstruction fidelity, even if its nominal BPP is slightly larger.

Given this, and for the sake of completeness, we report a weighted BPP (wBPP), which adjusts the conventional BPP by incorporating the diversity of unique RGB values in the decoded image:

\begin{equation}
\label{eq:wbpp}
\mathrm{wBPP} = \frac{\mathrm{BPP}}{\log_{2}(1 + N_{\text{unique}})},
\end{equation}
where $N_{\text{unique}}$ denotes the number of unique RGB triplets in the reconstructed image. Baseline JPEG often collapses tonal values due to quantization, yielding fewer unique triplets, whereas our method preserves finer tonal diversity. Table~\ref{tab:wBPP} reports wBPP on the S24 test set~\cite{S24} across different JPEG qualities (25, 50, 75, 95, 100). As shown, our method consistently achieves the best reconstruction accuracy with only a small wBPP overhead compared to competing approaches. Notably, wBPP reflects tonal diversity and reveals a smaller disparity between our method and baseline methods than the conventional BPP reported in the main paper.

\subsection{Fixed Bitrate (BPP)}
\label{supp-sec:fixed-bpp}
To further evaluate the proposed Raw-JPEG Adapter under realistic compression constraints, we report results at approximately fixed bitrates measured in BPP. Instead of comparing models at fixed JPEG quality levels, we perform per-image rate control to match a target BPP.

Specifically, for each test image, we conduct a binary search over the JPEG quality parameter to reach the desired target BPP. For the Raw-JPEG Adapter, at each candidate JPEG quality we select the model trained at the closest quality level and encode the pre-processed raw image accordingly. The final reconstruction corresponds to the JPEG quality that yields a total BPP (including the serialized pre-processing parameters stored in the COM segment) closest to the target value. For a fair comparison, the same rate-control procedure is applied to standard JPEG compression (without any raw pre-processing).

Table~\ref{bpp-table} reports PSNR, SSIM, and MS-SSIM for Raw-JPEG Adapter (with and without the DCT component) and standard JPEG compression at target BPP values of 0.5, 0.75, 1.0, and 1.25. As shown, Raw-JPEG Adapter consistently achieves higher reconstruction fidelity at closely matched achieved BPP compared to direct JPEG compression of raw data.

\input{tables/bpp_tab}

\input{tables/ablation}

\subsection{Ablation studies} 
\label{sec:ablations}
We performed a set of ablation studies to validate our design choices. We evaluated configurations including learning the gamma map only, LuT only, enabling the DCT component, adding the ECA channel attention block~\cite{wang2020eca}, using the JPEG simulator~\cite{xing21invertible} versus a simple quantization baseline (see Sec.~\ref{sec:diff-jpeg}), varying the LuT size, and removing individual loss terms. As shown in Table ~\ref{tab:ablation}, the full design described in the main paper achieves the best overall results.

\input{tables/jpeg2000}

\input{tables/results_supp_2}

\input{tables/results_srgb}

\begin{figure}[!h]
\centering
\includegraphics[width=0.65\linewidth]{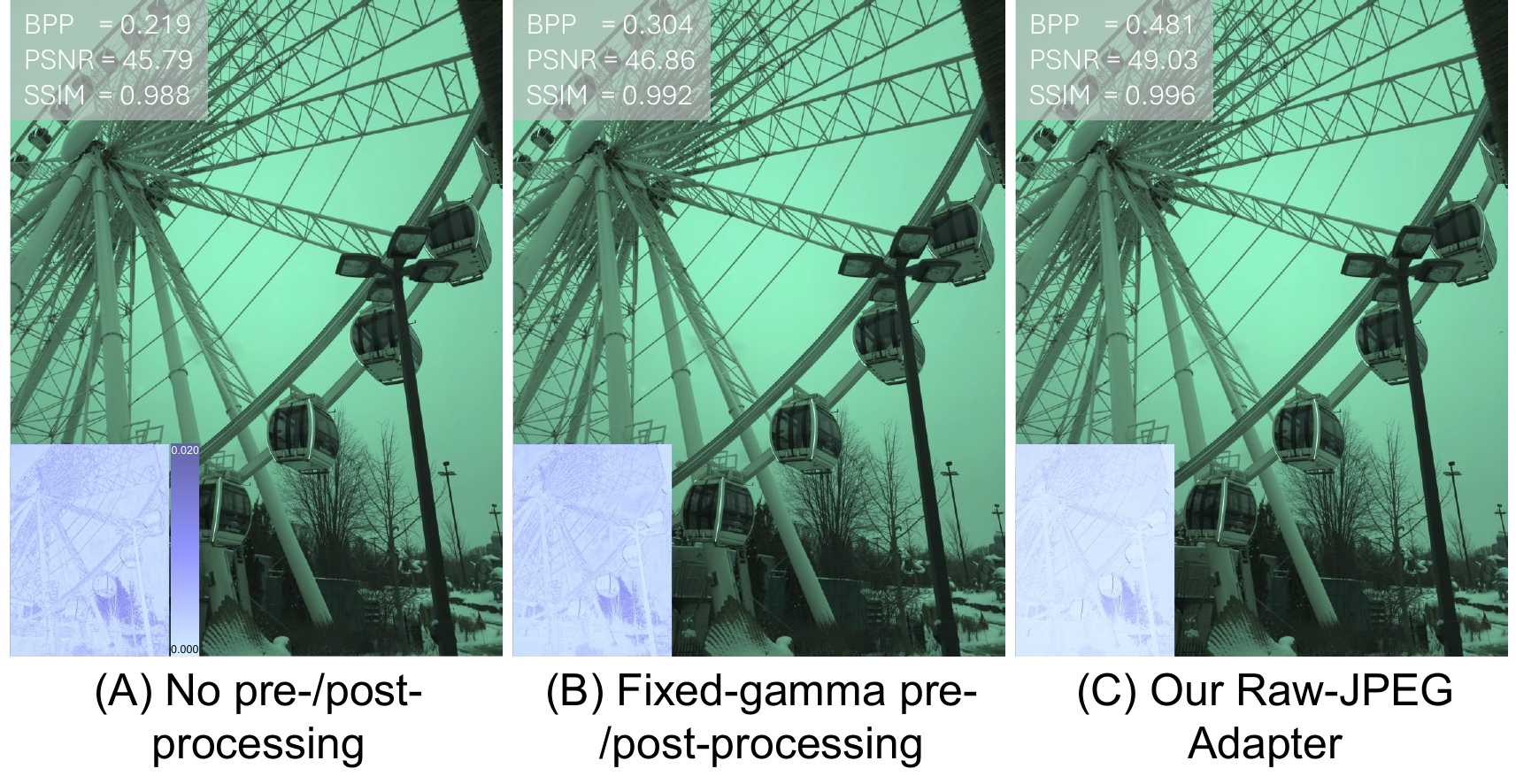}
\vspace{-2mm}
\caption{Qualitative example using neural compression instead of JPEG. We apply LIC-TCM~\cite{LIC} to compress: (A) the unprocessed raw image, (B) the raw image with fixed gamma pre-/post-processing, and (C) the raw image processed with our method. Error maps (w.r.t. the original raw) are shown in the lower-left corner of each image.\label{fig:neural_compression}}
\vspace{-3mm}
\end{figure}

\begin{figure}[!h]
\centering
\includegraphics[width=\linewidth]{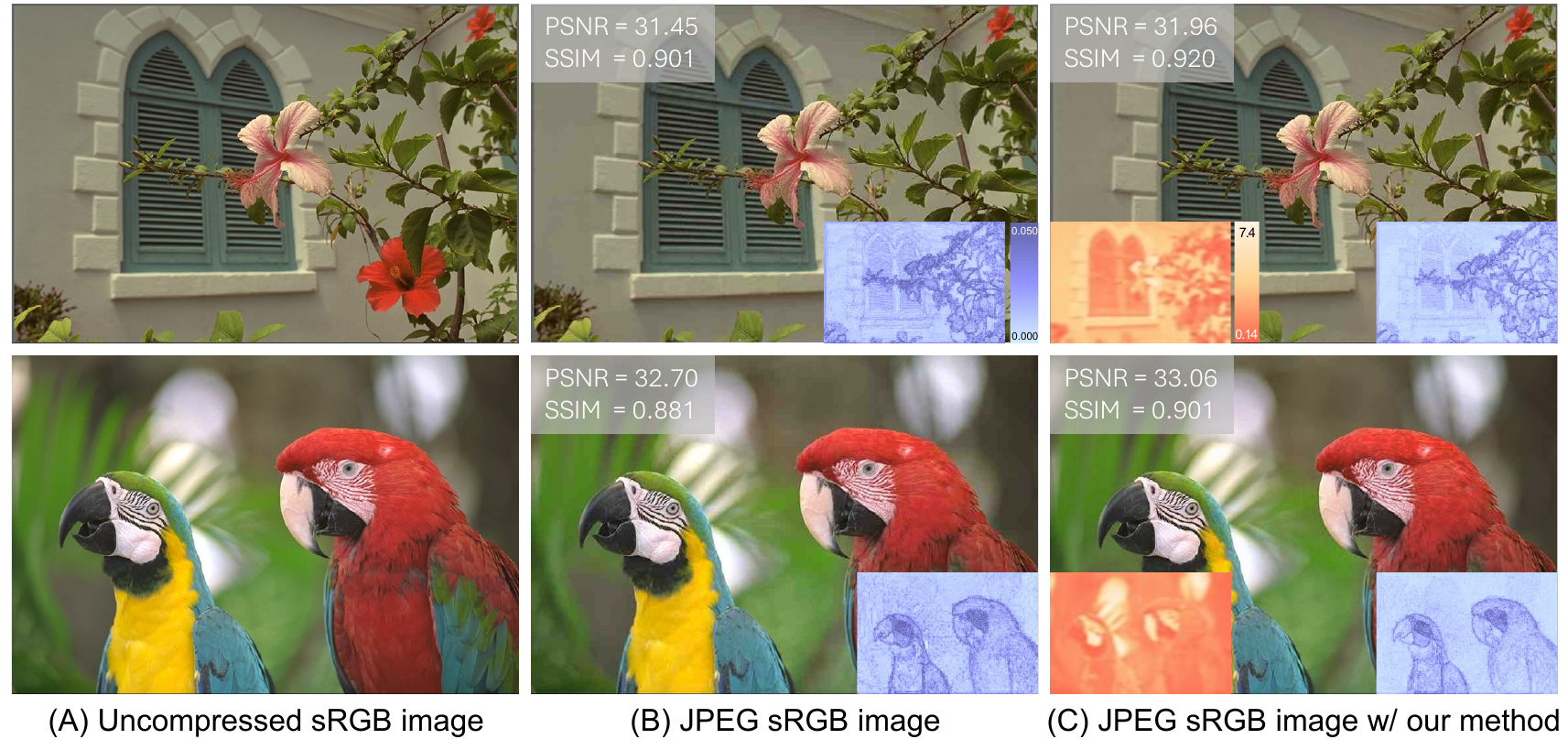}
\vspace{-4mm}
\caption{Qualitative examples of applying our method to sRGB JPEG image compression using the Kodak PhotoCD dataset \cite{Kodak1993} at JPEG quality 25. (A) Uncompressed demosaiced sRGB image. (B) JPEG-compressed sRGB image without pre-/post-processing, with its error map. (C) Decoded sRGB image produced by our method, along with the corresponding error map and predicted gamma map (in orange). \label{fig:srgb}}
\vspace{-3mm}
\end{figure}

\begin{figure*}[!h]
\centering
\includegraphics[width=\linewidth]{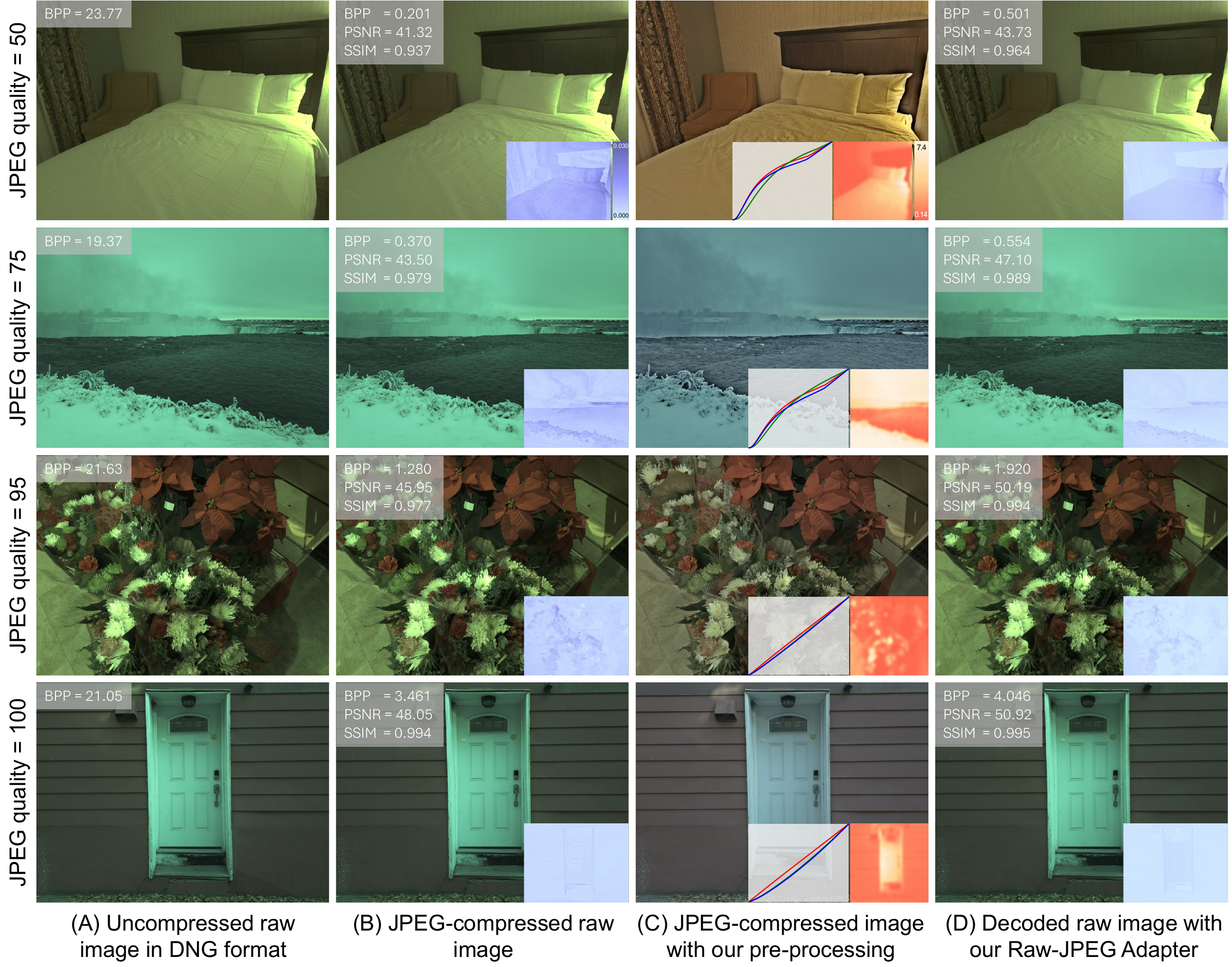}
\vspace{-4mm}
\caption{Qualitative examples from the S24 test set \cite{S24} at JPEG qualities 25, 50, 75, and 95. (A) Uncompressed demosaiced raw image. (B) JPEG-compressed raw without pre-/post-processing, with its error map. (C) JPEG image produced by our Raw-JPEG Adapter using the predicted RGB LUTs and gamma map. (D) Decoded raw image by our Raw-JPEG Adapter, along with the corresponding error map. \label{fig:qualitative_fig_0_supp}}
\vspace{-3mm}
\end{figure*}

\begin{figure*}[!h]
\centering
\includegraphics[width=\linewidth]{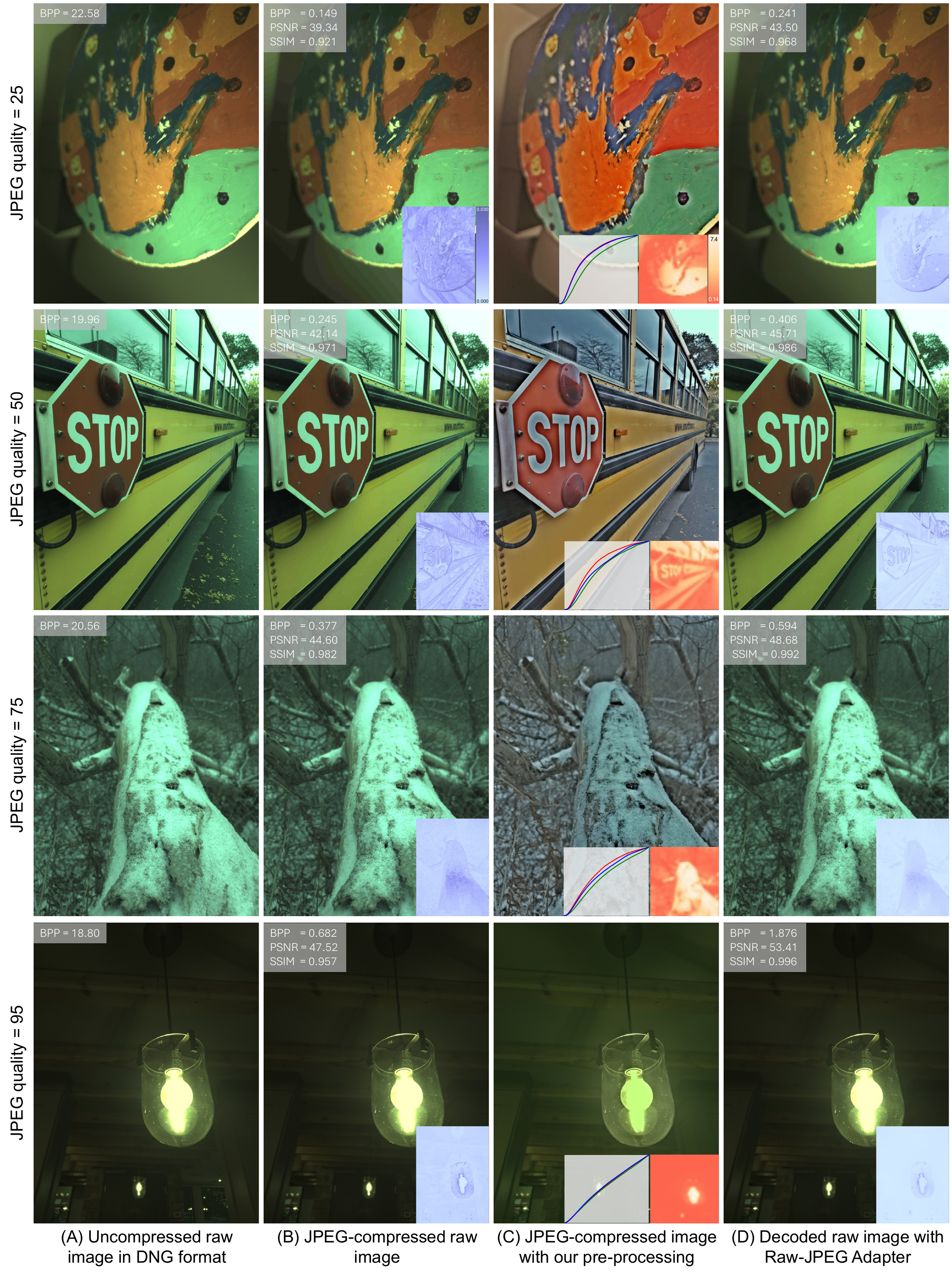}
\vspace{-4mm}
\caption{Additional qualitative examples from the S24 test set \cite{S24} at JPEG qualities 25, 50, 75, and 95. (A) Uncompressed demosaiced raw image. (B) JPEG-compressed raw without pre-/post-processing, with its error map. (C) JPEG image produced by our Raw-JPEG Adapter using the predicted RGB LUTs and gamma map. (D) Decoded raw image by our Raw-JPEG Adapter, along with the corresponding error map. \label{fig:qualtative_nus_75_supp}}
\vspace{-3mm}
\end{figure*}

\begin{figure*}[!h]
\centering
\includegraphics[width=\linewidth]{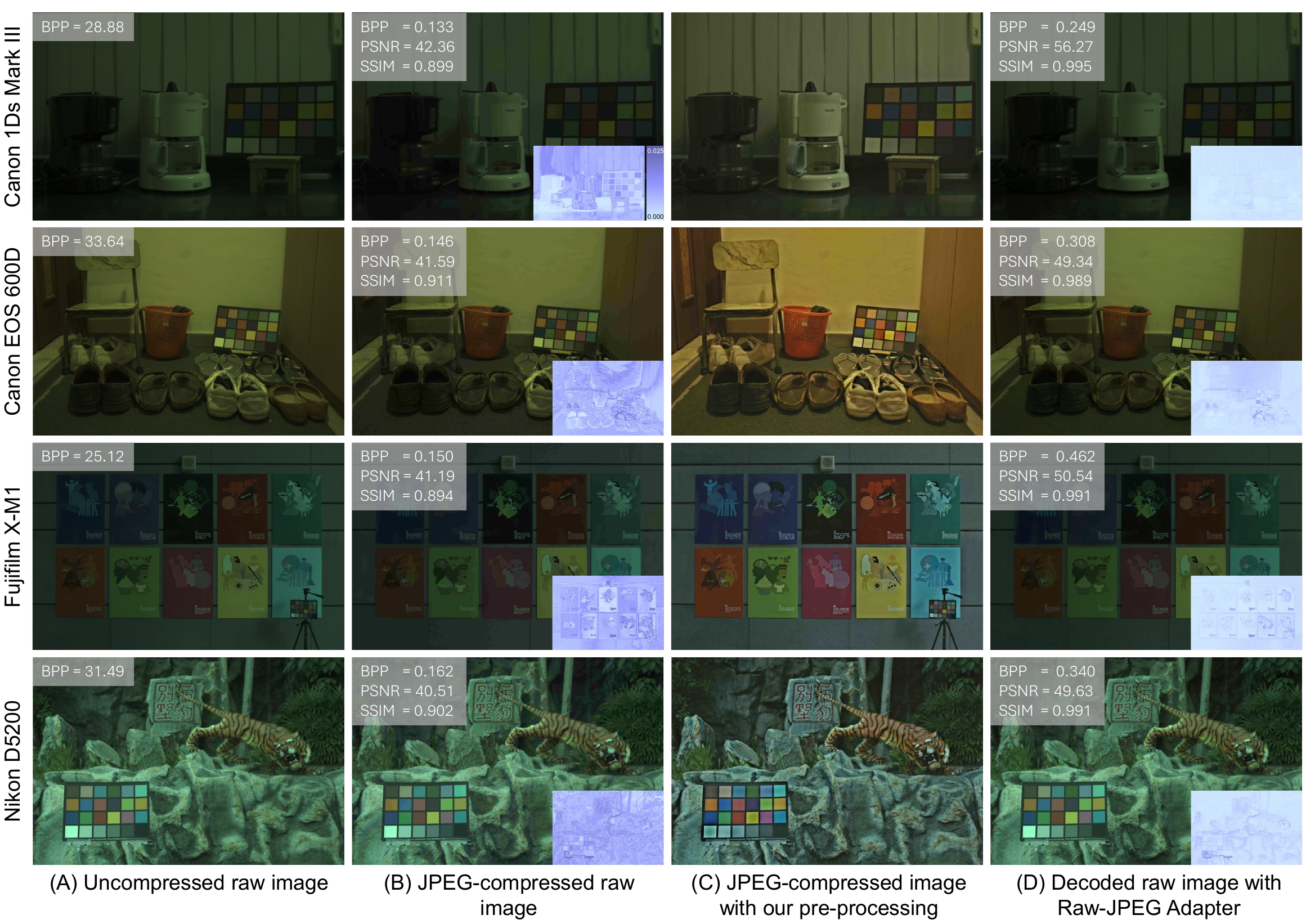}
\vspace{-4mm}
\caption{Qualitative examples from the NUS 8-camera dataset \cite{NUS} at JPEG quality 25. (A) Uncompressed demosaiced raw image. (B) JPEG-compressed raw image without pre-/post-processing, with its error map. (C) JPEG image produced by our Raw-JPEG Adapter. (D) Decoded raw image generated by our Raw-JPEG Adapter, along with the corresponding error map.  Note that our model was trained on the S24 dataset and has never seen examples from the NUS DSLR cameras. \label{fig:qualtative_nus_25_supp}}
\vspace{-3mm}
\end{figure*}

\begin{figure*}[!h]
\centering
\includegraphics[width=\linewidth]{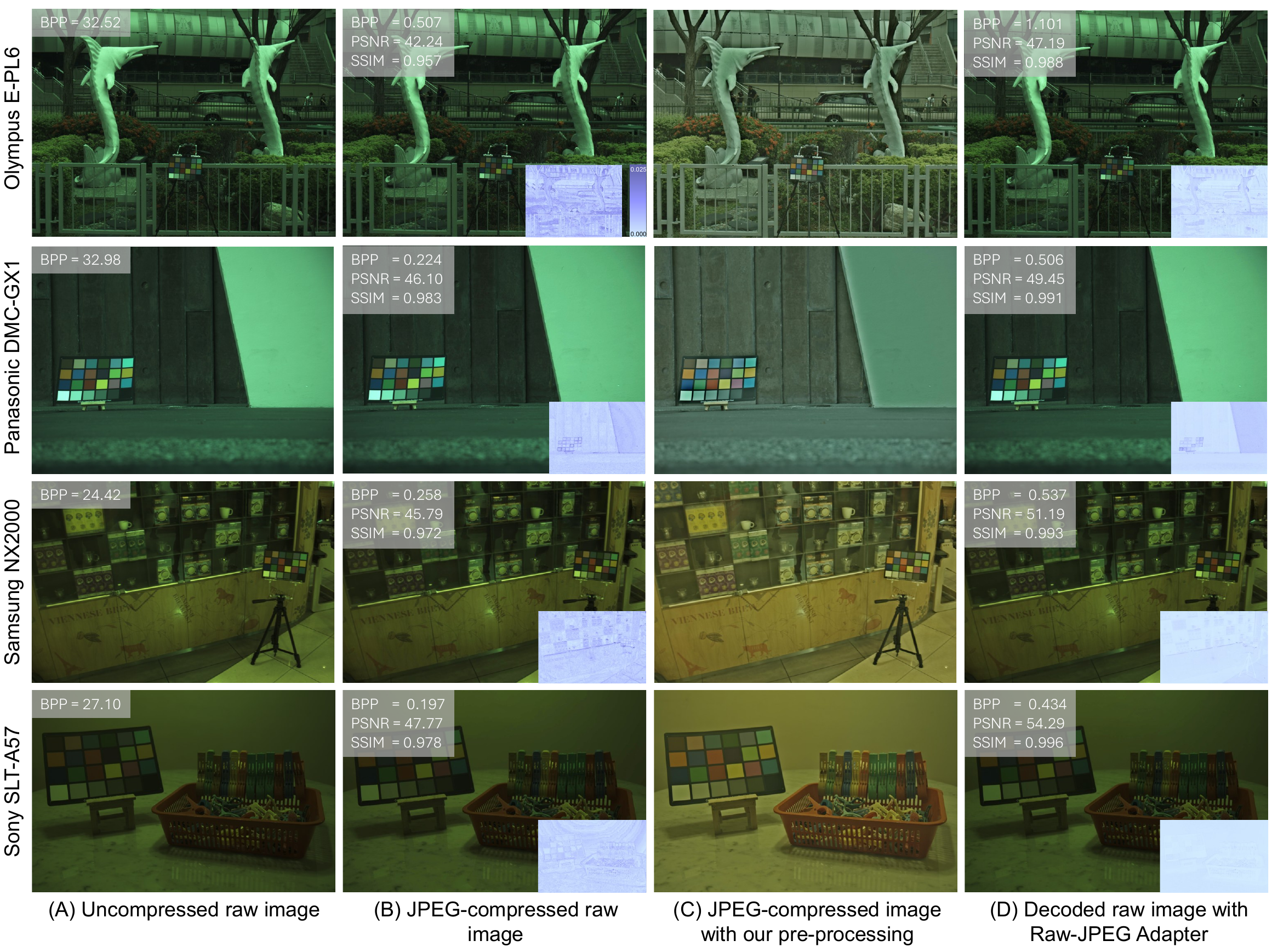}
\vspace{-4mm}
\caption{Qualitative examples from the NUS 8-camera dataset \cite{NUS} at JPEG quality 75. (A) Uncompressed demosaiced raw image. (B) JPEG-compressed raw image without pre-/post-processing, with its error map. (C) JPEG image produced by our Raw-JPEG Adapter. (D) Decoded raw image generated by our Raw-JPEG Adapter, along with the corresponding error map. Note that our model was trained on the S24 dataset and has never seen examples from the NUS DSLR cameras.  \label{fig:qualtative_nus_75_supp}}
\vspace{-4mm}
\end{figure*}

\begin{figure*}[!h]
\centering
\includegraphics[width=\linewidth]{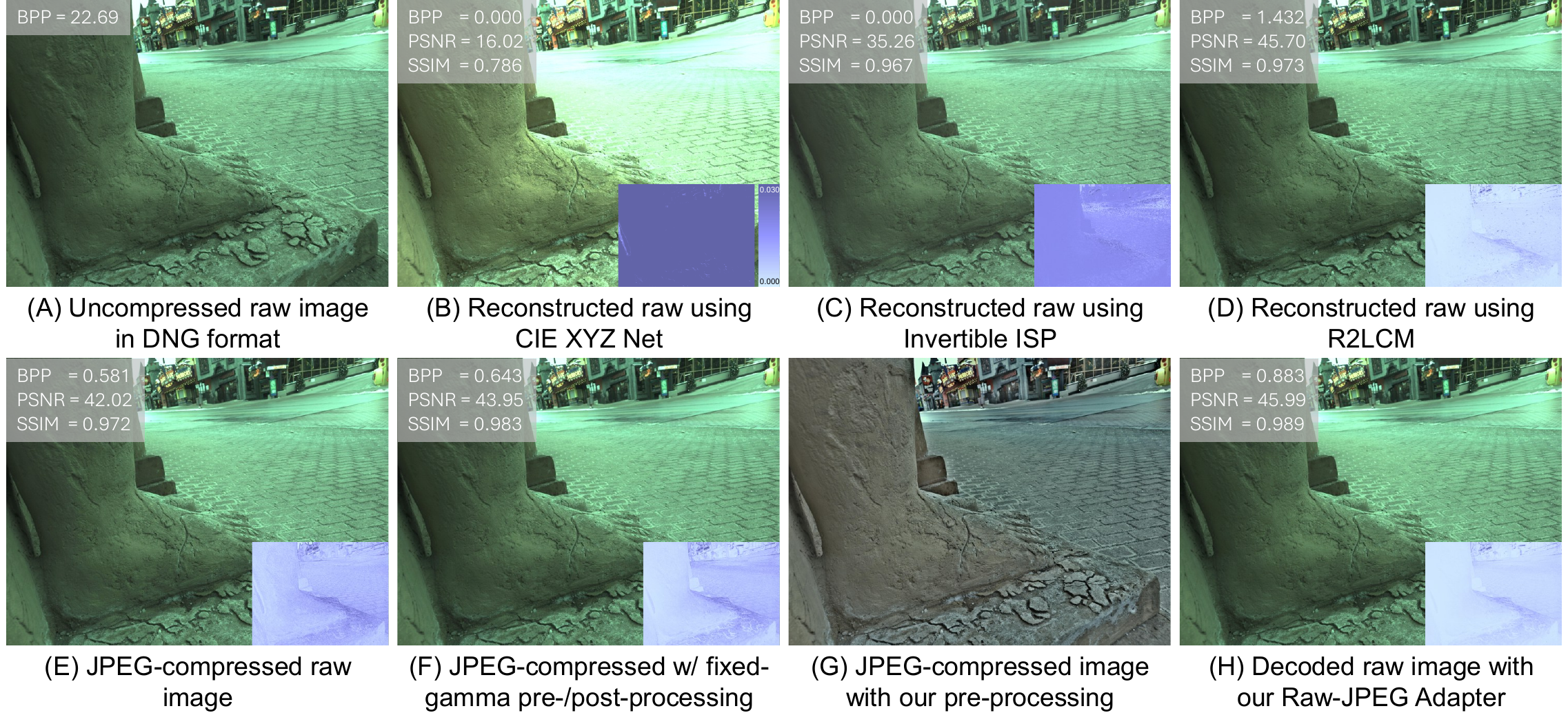}
\vspace{-4mm}
\caption{Qualitative comparison on the S24 test set~\cite{S24}. (A) Uncompressed demosaiced raw image. (B) Reconstructed raw image by CIE XYZ Net~\cite{afifi2021cie}. (C) Reconstructed raw image by Invertible ISP~\cite{xing21invertible}. (D) Reconstructed raw image by R2LCM~\cite{R2LCM1, R2LCM2}. (E) JPEG-compressed raw image without pre-/post-processing (JPEG quality 75). (F) Decoded raw image after fixed-gamma pre-/post-processing (JPEG quality 75). (G) JPEG image produced by our Raw-JPEG Adapter (JPEG quality 75). (H) Decoded raw image from our Raw-JPEG Adapter. The corresponding error maps are shown for (B–F) and (H). \label{fig:qualtative_2}}
\vspace{-1mm}
\end{figure*}

\begin{figure*}[!h]
\centering
\includegraphics[width=\linewidth]{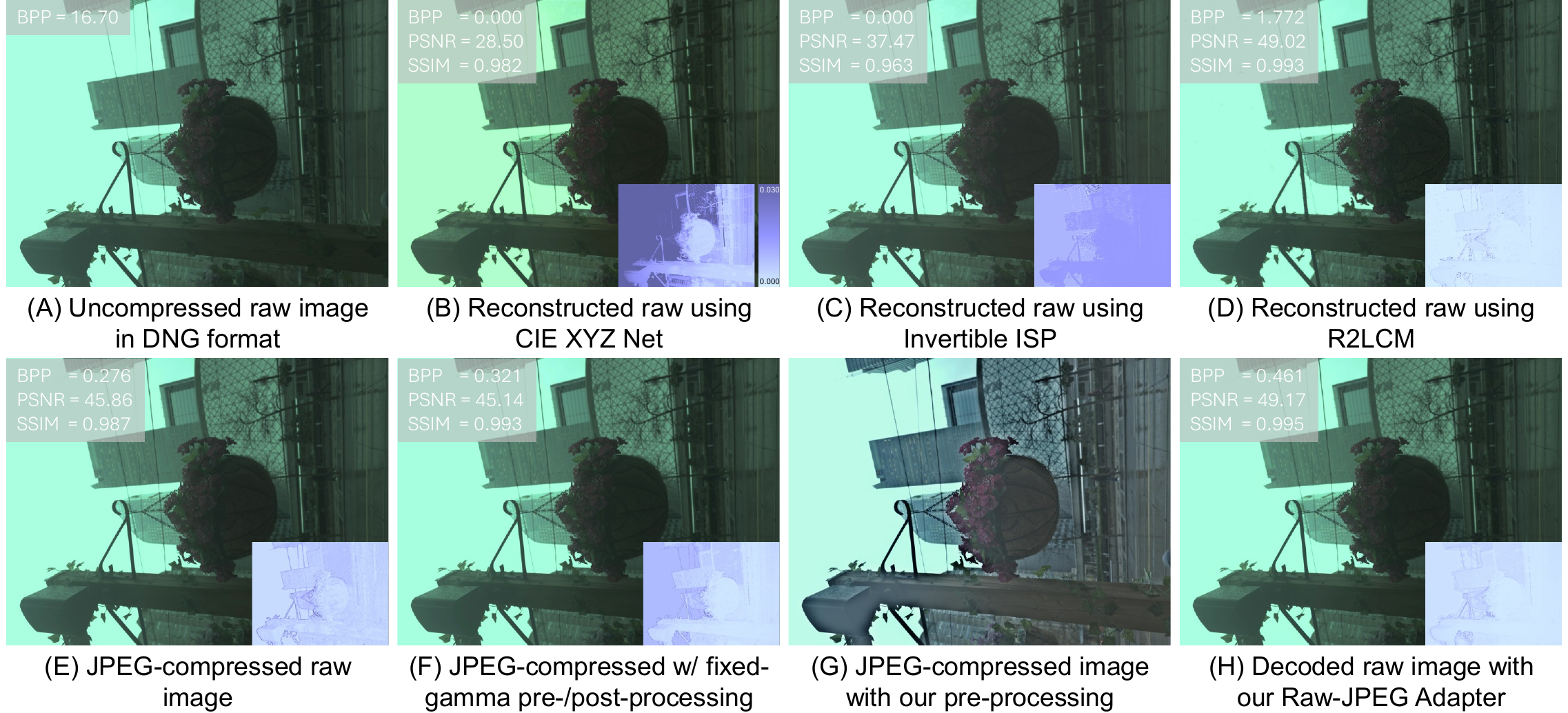}
\vspace{-4mm}
\caption{Additional qualitative comparison on the S24 test set~\cite{S24}. (A) Uncompressed demosaiced raw image. (B) Reconstructed raw image by CIE XYZ Net~\cite{afifi2021cie}. (C) Reconstructed raw image by Invertible ISP~\cite{xing21invertible}. (D) Reconstructed raw image by R2LCM~\cite{R2LCM1, R2LCM2}. (E) JPEG-compressed raw image without pre-/post-processing (JPEG quality 75). (F) Decoded raw image after fixed-gamma pre-/post-processing (JPEG quality 75). (G) JPEG image produced by our Raw-JPEG Adapter (JPEG quality 75). (H) Decoded raw image from our Raw-JPEG Adapter. The corresponding error maps are shown for (B–F) and (H). \label{fig:qualtative_2_supp}}
\vspace{-1mm}
\end{figure*}

\begin{figure*}[!h]
\centering
\includegraphics[width=\linewidth]{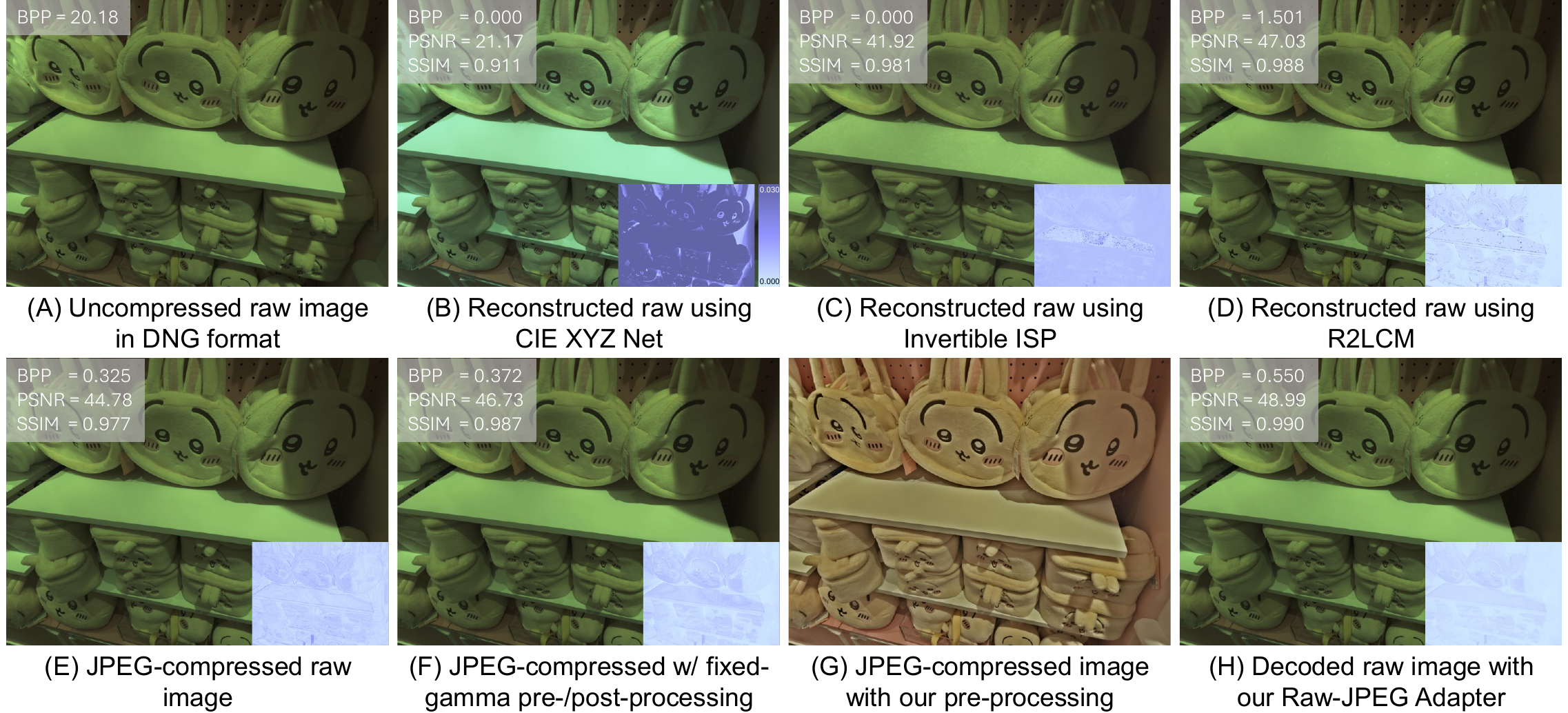}
\vspace{-4mm}
\caption{Additional qualitative comparison on the S24 test set~\cite{S24}. (A) Uncompressed demosaiced raw image. (B) Reconstructed raw image by CIE XYZ Net~\cite{afifi2021cie}. (C) Reconstructed raw image by Invertible ISP~\cite{xing21invertible}. (D) Reconstructed raw image by R2LCM~\cite{R2LCM1, R2LCM2}. (E) JPEG-compressed raw image without pre-/post-processing (JPEG quality 75). (F) Decoded raw image after fixed-gamma pre-/post-processing (JPEG quality 75). (G) JPEG image produced by our Raw-JPEG Adapter (JPEG quality 75). (H) Decoded raw image from our Raw-JPEG Adapter. The corresponding error maps are shown for (B–F) and (H). \label{fig:qualtative_3_supp}}
\vspace{-1mm}
\end{figure*}

\subsection{Generalization}
While our method is primarily designed for the JPEG codec, the underlying framework is not restricted to JPEG. In particular, the pre-processing and decoding pipeline can be paired with other compression standards or learned image codecs. As demonstrated in the main paper, we reported results of using our pipeline to improve raw compression accuracy with the deep image compression method LIC-TCM~\cite{LIC}. Specifically, we showed that applying our method as a pre-processing step before compressing the raw image with LIC-TCM yields better results compared to using LIC-TCM directly or applying a fixed gamma. A qualitative example is shown in Fig.~\ref{fig:neural_compression}. We further evaluated our trained models at inference by replacing the JPEG codec with JPEG~2000~\cite{marcellin2000overview}, using models trained with quality levels 50 and 75 in the JPEG simulator. As shown in Table~\ref{tab:jpeg-2000}, our method consistently improved reconstruction accuracy in this setting as well.

Our method, combined with color and intensity augmentations, demonstrates strong generalization across datasets captured by cameras different from the one used during training, as shown in the main paper. There, we reported results under various JPEG qualities (25, 50, 75, and 95). Here, we further evaluate our method at the highest JPEG quality setting ($Q=100$) across these datasets from unseen cameras; see Table~\ref{tab:cross-camera-quality-100-supp}.

While our method is designed for raw image compression, we also explored its applicability to sRGB image compression. To this end, we generated uncompressed sRGB images from the S24 training set~\cite{S24} by re-rendering the raw data with Adobe Photoshop (since the provided sRGB images in the S24 dataset are already JPEG-compressed). We then trained our model in a self-supervised manner on these uncompressed sRGB images and evaluated it on the Kodak PhotoCD dataset~\cite{Kodak1993}, a widely used benchmark for sRGB image compression. As shown in Table~\ref{tab:srgb} and Fig.~\ref{fig:srgb}, our method achieves slight but consistent improvements in reconstruction accuracy over standard JPEG, demonstrating its potential to generalize beyond raw image compression.

\begin{figure*}[!h]
\centering
\includegraphics[width=\linewidth]{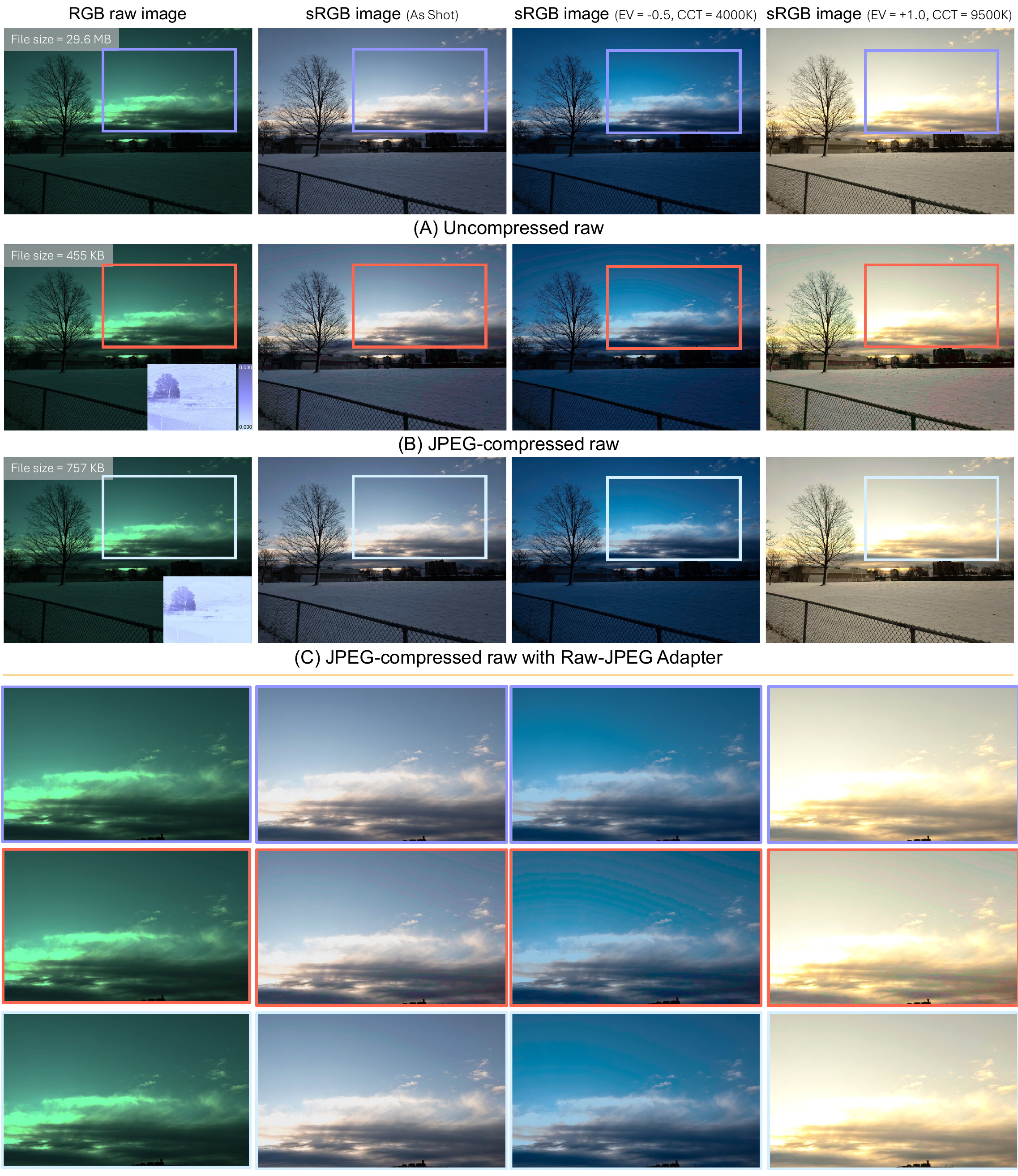}
\vspace{-5mm}
\caption{One key advantage of saving raw images is enabling flexible post-capture editing. This figure shows sRGB outputs from Adobe Lightroom after applying different edits to: (A) the original uncompressed raw image, (B) the JPEG-compressed raw image, and (C) the JPEG-compressed raw image processed with our Raw-JPEG Adapter. For (B) and (C), JPEG files were saved with quality set to 50. Error maps between the original and decoded raw images are also shown. Our method enables high-quality raw editing while significantly reducing file size. \label{fig:lightroom_1}}
\vspace{-4mm}
\end{figure*}

\begin{figure*}[!h]
\centering
\includegraphics[width=\linewidth]{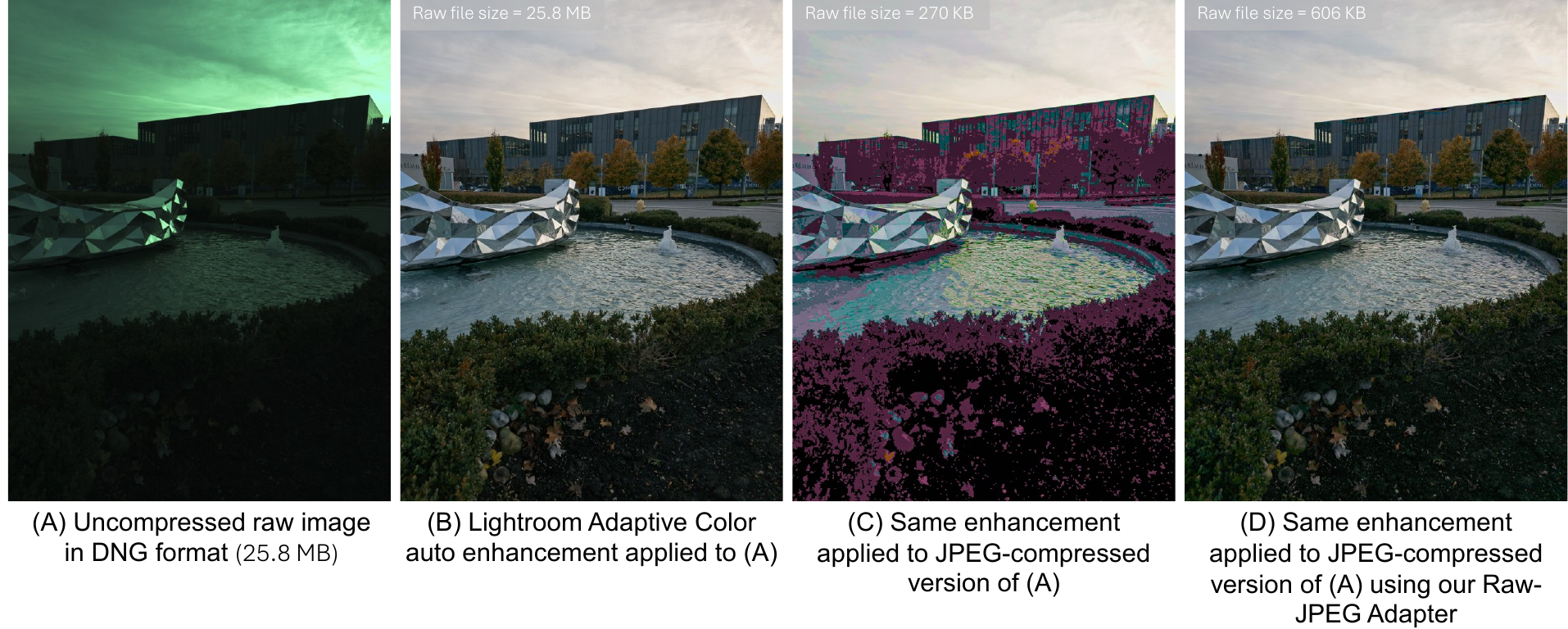}
\vspace{-5mm}
\caption{Additional example of editing using Adobe Lightroom. This figure shows automatic enhancement of a raw image using the Adaptive Color Profile in Adobe Lightroom. We show: (B) the result of applying the enhancement to the original uncompressed raw image shown in (A), (C) the result using a JPEG-compressed raw image with quality set to 50, and (D) the result using JPEG compression with our method, also at quality 50. \label{fig:lightroom_fig_2}}
\vspace{-1mm}
\end{figure*}

\subsection{Additional qualitative results}  
We show additional qualitative examples from the S24 test set~\cite{S24} and the NUS dataset~\cite{NUS}, using models trained primarily on the S24 training set~\cite{S24}, in Figs.~\ref{fig:qualitative_fig_0_supp}--\ref{fig:qualtative_nus_75_supp}. Further qualitative comparisons with other methods are provided in Figs.~\ref{fig:qualtative_2}--\ref{fig:qualtative_3_supp}.

\begin{figure*}[!h]
\centering
\includegraphics[width=\linewidth]{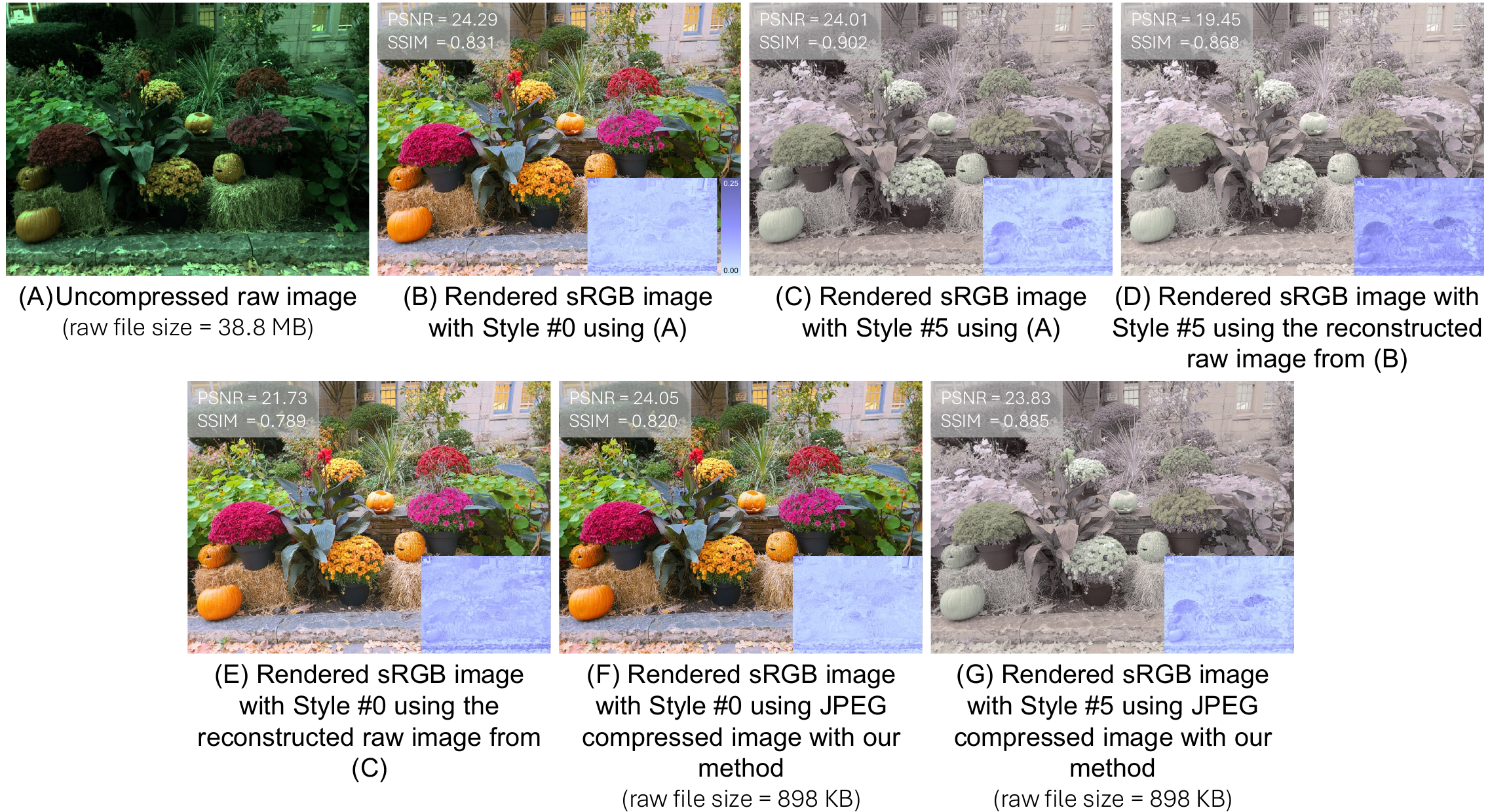}
\vspace{-5mm}
\caption{Invertible ISP~\cite{xing21invertible} image rendering and re-rendering across picture styles. (A) shows an uncompressed raw image rendered to two different styles from the S24 dataset~\cite{S24} -- Style \#0 and Style \#5 -- shown in (B) and (C). These rendered images are then used to reconstruct the raw image and re-rendered to the alternate style in (D) and (E). Our method enables storing a highly compressed raw representation (with a significantly smaller file size than the original uncompressed raw DNG file) that allows for direct re-rendering to different styles without the need to reconstruct the raw from sRGB, as demonstrated in (F) and (G) with higher accuracy. Error maps between the rendered sRGB and ground-truth sRGB images are shown in the lower-right corner of (B-G).
 \label{fig:re-rendering-1}}
\vspace{-3mm}
\end{figure*}

\subsection{Post-capture image re-rendering}

\input{tables/neural_isp_tab_1}

\input{tables/neural_isp_tab_2}

In the main paper, we showed that our method enables storing raw images in a significantly smaller file size while maintaining high fidelity. Unlike directly saving raw images with JPEG---which often introduces banding and quantization artifacts---our method preserves the raw signal with substantially less loss of detail and fewer artifacts. Consequently, the final display-referred sRGB images rendered from our saved raw exhibit perceptually fewer degradations compared to those saved directly as JPEG, while remaining comparable in quality to those rendered from the original DNG files, which require substantially larger storage.

We further provide qualitative comparisons in Fig.~\ref{fig:lightroom_1}. In this figure, we render our JPEG-compressed raw images using Adobe Lightroom under different rendering settings and compare them against raw images saved directly as JPEG and against the original uncompressed raw. In Fig.~\ref{fig:lightroom_fig_2}, we show results with Adobe Lightroom’s automatic enhancement Adaptive Color Profile, comparing our method with direct JPEG saving and the uncompressed raw. As can be seen, our method produces sRGB images that are visually similar to those obtained from the uncompressed raw, while requiring a significantly smaller file size. In contrast, directly saving raw images as JPEG without our method consistently introduces color shifts due to quantization and visible banding artifacts.

We complement these visual results with quantitative evaluation by rendering raw images through two neural ISPs, LiteISP~\cite{lite-isp} and Invertible ISP~\cite{xing21invertible}, both trained to map raw images from the Samsung S24 Ultra main camera’s raw space to the sRGB space under the default picture style of the S24 dataset~\cite{S24}. We report the accuracy of the rendered sRGB images produced by the two neural ISPs against the ground-truth sRGB images. The evaluation includes: 1) uncompressed raw images (from the provided 16-bit PNG raw files in the S24 dataset~\cite{S24}), 2) JPEG-compressed raw images obtained by saving the 16-bit PNG raw files as 8-bit JPEGs without pre-/post-processing, 3) JPEG raw images saved with fixed gamma (applying a 2.2 gamma correction before saving and inverting it after decoding), and 4) our method (with and without the DCT component). The results are shown in Table~\ref{tab:neural-isp-1}, where it can be seen that our method consistently achieves the best performance compared to other JPEG-based raw storage approaches across different JPEG qualities.

We further investigate the ability of our method to save raw images in an affordable file size while preserving the flexibility of re-rendering under different picture styles. Specifically, after rendering an image from our JPEG-compressed raw, the raw can be re-rendered multiple times (e.g., interactively) without the need to store a large DNG file. In comparison, bidirectional neural ISPs such as the Invertible ISP~\cite{xing21invertible} also enable this functionality, where a rendered sRGB image can be mapped back to raw and subsequently re-rendered.

We compare these two scenarios: 1) storing the raw image with our JPEG-based method at a small file size and re-rendering directly, versus 2) reconstructing raw from sRGB images rendered by the Invertible ISP~\cite{xing21invertible}. For the latter, we reconstruct raw images from sRGB renders produced from the original raw with the default style (picture style \#0) and style \#5 in the S24 dataset~\cite{S24}, and then re-render them into different target picture styles (\#1, \#2, \#3, …). Our method is evaluated under the same setting, using JPEG-compressed raw saved with our pipeline and repeatedly re-rendered.

The results are reported in Table~\ref{tab:neural-isp-2}. As shown, our method achieves higher overall accuracy while also being significantly more efficient (see Fig.~\ref{fig:re-rendering-1}). In particular, our JPEG-based raw representation requires only lightweight pre-processing and storage, with an average processing time of 0.12 seconds per image. In contrast, the Invertible ISP incurs a large computational overhead, taking on average 7.9 seconds per image to reconstruct the raw before re-rendering.

%% file: tables/wBPP.tex
\begin{table*}[t]
\centering
\caption{Quantitative results on the S24 test set~\cite{S24} across different JPEG quality levels, with wBPP (Eq.~\ref{eq:wbpp}) reported. The best results are highlighted in \textcolor{green}{\textbf{green}}, and the second-best are shown in \textbf{bold}. \label{tab:wBPP}}
\vspace{-2mm}

\scalebox{0.48}{
\centering
\begin{tabular}{|l|c|c|c|c|c|c|c|c|c|c|c|c|c|c|c|c|c|c|c|c|}
\hline
\multirow{2}{*}{\textbf{Method}} &
  \multicolumn{4}{c|}{\cellcolor[HTML]{fed18d}\textbf{Quality = 25}} &
  \multicolumn{4}{c|}{\cellcolor[HTML]{ff9484}\textbf{Quality = 50}} &
  \multicolumn{4}{c|}{\cellcolor[HTML]{d5effe}\textbf{Quality = 75}} &
  \multicolumn{4}{c|}{\cellcolor[HTML]{CBCEFB}\textbf{Quality = 95}} &
  \multicolumn{4}{c|}{\cellcolor[HTML]{fff0dd}\textbf{Quality = 100}} \\
\cline{2-21}
& \textbf{PSNR} & \textbf{\begin{tabular}[c]{@{}c@{}}SSIM\\ ($\times$100)\end{tabular}} & 
  \textbf{\begin{tabular}[c]{@{}c@{}}MS-SSIM\\ ($\times$100)\end{tabular}} & \textbf{wBPP} &
  \textbf{PSNR} & \textbf{\begin{tabular}[c]{@{}c@{}}SSIM\\ ($\times$100)\end{tabular}} &
  \textbf{\begin{tabular}[c]{@{}c@{}}MS-SSIM\\ ($\times$100)\end{tabular}} & \textbf{wBPP} &
  \textbf{PSNR} & \textbf{\begin{tabular}[c]{@{}c@{}}SSIM\\ ($\times$100)\end{tabular}} &
  \textbf{\begin{tabular}[c]{@{}c@{}}MS-SSIM\\ ($\times$100)\end{tabular}} & \textbf{wBPP} &
  \textbf{PSNR} & \textbf{\begin{tabular}[c]{@{}c@{}}SSIM\\ ($\times$100)\end{tabular}} &
  \textbf{\begin{tabular}[c]{@{}c@{}}MS-SSIM\\ ($\times$100)\end{tabular}} & \textbf{wBPP} &
  \textbf{PSNR} & \textbf{\begin{tabular}[c]{@{}c@{}}SSIM\\ ($\times$100)\end{tabular}} &
  \textbf{\begin{tabular}[c]{@{}c@{}}MS-SSIM\\ ($\times$100)\end{tabular}} & \textbf{wBPP} \\
\hline

JPEG &
  36.58 & 90.08 & 95.79 & 0.02 & 
  39.02 & 93.57 & 97.74 & 0.02 & 
  40.75 & 95.40 & 98.62 & 0.03 & 
  43.60 & 97.21 & 99.35 & 0.11 & 
  46.00 & 98.47 & 99.72 & 0.28 \\ \hline

JPEG + raw$\leftrightarrow$sRGB &
  39.41 & 95.25 & 98.39 & 0.02 & 
  41.26 & 96.58 & 99.07 & 0.03 & 
  42.63 & 97.37 & 99.39 & 0.03 & 
  45.42 & 98.52 & 99.62 & 0.16 & 
  47.65 & 99.25 & 99.86 &  0.38 \\ \hline

JPEG + fixed gamma &
  38.30 & 94.51 & 97.82 & 0.02 & 
  40.74 & 96.20 & 98.81 & 0.03 & 
  42.43 & 97.16 & 99.25 & 0.05 & 
  45.42 & 98.42 & 99.65 & 0.14 & 
  48.38 & 99.27 & \textbf{99.87} & 0.34 \\ \hdashline

Ours (w/o DCT, w/o aug) &
  39.35 & 95.71 & 98.69 & 0.03 & 
  41.69 & 96.91 & 99.22 & 0.04 & 
  43.31 & 97.61 & 99.46 & 0.06 & 
  45.97 & 98.58 & 99.71 & 0.14 & 
  48.49 & \textbf{99.28} & \textbf{99.87} &  0.35 \\ \hline

Ours (w/o DCT, w/ aug) &
  39.49 & 95.70 & 98.69 & 0.03 & 
  41.73 & 96.87 & 99.20 &  0.04 & 
  43.36 & 97.60 & 99.46 & 0.06 & 
  46.12 & \cellcolor{best}\textbf{98.61} & \cellcolor{best}\textbf{99.72} & 0.14 & 
  48.44 & \textbf{99.28} & \textbf{99.87} & 0.35 \\ \hline

Ours (w/ DCT, w/o aug) &
  \textbf{39.70} & \textbf{95.83} & \textbf{98.77} & 0.03 & 
  \cellcolor{best}\textbf{41.92} & \cellcolor{best}\textbf{96.99} & \cellcolor{best}\textbf{99.26} & 0.04 & 
  \textbf{43.54} & \cellcolor{best}\textbf{97.68} & \cellcolor{best}\textbf{99.49} & 0.06 & 
  \cellcolor{best}\textbf{46.22} & \textbf{98.60} & \cellcolor{best}\textbf{99.72} & 0.14 & 
  \textbf{48.99} & \cellcolor{best}\textbf{99.35} & \cellcolor{best}\textbf{99.89} & 0.35 \\ \hline

Ours (w/ DCT, w/ aug) &
  \cellcolor{best}\textbf{39.82} & \cellcolor{best}\textbf{95.88} & \cellcolor{best}\textbf{98.78} & 0.03 & 
  \textbf{41.88} & \textbf{96.98} & \textbf{99.25} & 0.04 & 
  \cellcolor{best}\textbf{43.58} & \textbf{97.67} & \textbf{99.48} & 0.06 & 
  \textbf{46.19} & 98.58 & \textbf{99.71} & 0.14 & 
  \cellcolor{best}\textbf{49.04} & \cellcolor{best}\textbf{99.35} & \cellcolor{best}\textbf{99.89} & 0.35 \\ \hline

\end{tabular}
}\vspace{-2mm}
\end{table*}

%% file: tables/bpp_tab.tex
\begin{table*}[t]
\centering
\caption{Quantitative results on the S24 test set \cite{S24} at different target bit-per-pixel (BPP) levels. For each method, we report the actual achieved BPP, along with PSNR, SSIM ($\times$100), and MS-SSIM ($\times$100). Best results are highlighted in \textcolor{green}{\textbf{green}}, and second-best in \textbf{bold}.\label{bpp-table}}
\vspace{-2mm}

\scalebox{0.58}{
\begin{tabular}{|l|cccc|cccc|cccc|cccc|}
\hline
\multirow{2}{*}{\textbf{Method}} &
  \multicolumn{4}{c|}{\cellcolor[HTML]{fed18d}\textbf{BPP = 0.50}} &
  \multicolumn{4}{c|}{\cellcolor[HTML]{ff9484}\textbf{BPP = 0.75}} &
  \multicolumn{4}{c|}{\cellcolor[HTML]{d5effe}\textbf{BPP = 1.00}} &
  \multicolumn{4}{c|}{\cellcolor[HTML]{CBCEFB}\textbf{BPP = 1.25}} \\
\cline{2-17}

& 
  \multicolumn{1}{c|}{\textbf{PSNR}} &
  \multicolumn{1}{c|}{\textbf{\begin{tabular}[c]{@{}c@{}}SSIM\\ ($\times$100)\end{tabular}}} &
  \multicolumn{1}{c|}{\textbf{\begin{tabular}[c]{@{}c@{}}MS-SSIM\\ ($\times$100)\end{tabular}}} &
  \multicolumn{1}{c|}{\textbf{BPP}} &
  \multicolumn{1}{c|}{\textbf{PSNR}} &
  \multicolumn{1}{c|}{\textbf{\begin{tabular}[c]{@{}c@{}}SSIM\\ ($\times$100)\end{tabular}}} &
  \multicolumn{1}{c|}{\textbf{\begin{tabular}[c]{@{}c@{}}MS-SSIM\\ ($\times$100)\end{tabular}}} &
  \multicolumn{1}{c|}{\textbf{BPP}} &
  \multicolumn{1}{c|}{\textbf{PSNR}} &
  \multicolumn{1}{c|}{\textbf{\begin{tabular}[c]{@{}c@{}}SSIM\\ ($\times$100)\end{tabular}}} &
  \multicolumn{1}{c|}{\textbf{\begin{tabular}[c]{@{}c@{}}MS-SSIM\\ ($\times$100)\end{tabular}}} &
    \multicolumn{1}{c|}{\textbf{BPP}} &
  \multicolumn{1}{c|}{\textbf{PSNR}} &
  \multicolumn{1}{c|}{\textbf{\begin{tabular}[c]{@{}c@{}}SSIM\\ ($\times$100)\end{tabular}}} &
  \multicolumn{1}{c|}{\textbf{\begin{tabular}[c]{@{}c@{}}MS-SSIM\\ ($\times$100)\end{tabular}}} &
  \multicolumn{1}{c|}{\textbf{BPP}} \\
\hline

JPEG &
  \multicolumn{1}{c|}{38.55} & \multicolumn{1}{c|}{91.47} & \multicolumn{1}{c|}{96.11} & \multicolumn{1}{c|}{0.50} &
  \multicolumn{1}{c|}{39.97} & \multicolumn{1}{c|}{93.58} & \multicolumn{1}{c|}{97.47} & \multicolumn{1}{c|}{0.75} &
  \multicolumn{1}{c|}{40.69} & \multicolumn{1}{c|}{94.26} & \multicolumn{1}{c|}{97.89} & \multicolumn{1}{c|}{0.99} &
  \multicolumn{1}{c|}{40.83} & \multicolumn{1}{c|}{93.67} & \multicolumn{1}{c|}{97.27} & \multicolumn{1}{c|}{1.20} \\
\hdashline

Raw-JPEG Adapter (w/o DCT) &
  \multicolumn{1}{c|}{\textbf{40.26}} & \multicolumn{1}{c|}{\textbf{95.62}} & \multicolumn{1}{c|}{\textbf{98.67}} & \multicolumn{1}{c|}{0.50} &
  \multicolumn{1}{c|}{\textbf{42.20}} & \multicolumn{1}{c|}{\textbf{96.85}} & \multicolumn{1}{c|}{\textbf{99.19}} & \multicolumn{1}{c|}{0.75} &
  \multicolumn{1}{c|}{\textbf{43.28}} & \multicolumn{1}{c|}{\textbf{97.31}} & \multicolumn{1}{c|}{\textbf{99.31}} & \multicolumn{1}{c|}{1.00} &
  \multicolumn{1}{c|}{\textbf{44.14}} & \multicolumn{1}{c|}{\textbf{97.76}} & \multicolumn{1}{c|}{\textbf{99.50}} & \multicolumn{1}{c|}{1.25} \\\hline

Raw-JPEG Adapter (w/$\text{ }$ DCT) &
  \multicolumn{1}{c|}{\cellcolor{best}\textbf{40.44}} & \multicolumn{1}{c|}{\cellcolor{best}\textbf{95.72}} & \multicolumn{1}{c|}{\cellcolor{best}\textbf{98.72}} & \multicolumn{1}{c|}{0.50} &
  \multicolumn{1}{c|}{\cellcolor{best}\textbf{42.35}} & \multicolumn{1}{c|}{\cellcolor{best}\textbf{96.91}} & \multicolumn{1}{c|}{\cellcolor{best}\textbf{99.22}} & \multicolumn{1}{c|}{0.75} &
  \multicolumn{1}{c|}{\cellcolor{best}\textbf{43.44}} & \multicolumn{1}{c|}{\cellcolor{best}\textbf{97.45}} & \multicolumn{1}{c|}{\cellcolor{best}\textbf{99.41}} & \multicolumn{1}{c|}{1.00} &
  \multicolumn{1}{c|}{\cellcolor{best}\textbf{44.25}} & \multicolumn{1}{c|}{\cellcolor{best}\textbf{97.78}} & \multicolumn{1}{c|}{\cellcolor{best}\textbf{99.51}} & \multicolumn{1}{c|}{1.24} \\
\hline

\end{tabular}
}
\vspace{-1mm}
\end{table*}

%% file: tables/ablation.tex
\begin{table}
\centering
\caption{Ablation study results for different variations of our proposed design, evaluated on the S24 test set \cite{S24} with JPEG quality set to 75. Best results are highlighted in \textcolor{green}{\textbf{green}}. \label{tab:ablation}\vspace{-1mm}}
\scalebox{0.65}{
\begin{tabular}{|ccccccccccc|c|c|}
\hline
\textbf{\begin{tabular}[c]{@{}c@{}}Gamma\\ (100$\times$100)\end{tabular}} &
  \textbf{\begin{tabular}[c]{@{}c@{}}LuT\\ (128$\times$3)\end{tabular}} &
  \textbf{\begin{tabular}[c]{@{}c@{}}DCT\end{tabular}} &
  \textbf{ECA} &
  \textbf{\begin{tabular}[c]{@{}c@{}}JPEG\\ sim.\end{tabular}} &
  \textbf{\begin{tabular}[c]{@{}c@{}}Gamma \\ (64$\times$64)\end{tabular}} &
  \textbf{\begin{tabular}[c]{@{}c@{}}LuT\\ (64$\times$3)\end{tabular}} &
  \textbf{\begin{tabular}[c]{@{}c@{}}LuT\\ (128$\times$1)\end{tabular}} &
  \textbf{\begin{tabular}[c]{@{}c@{}}L1\\ loss\end{tabular}} &
  \textbf{\begin{tabular}[c]{@{}c@{}}SSIM\\ loss\end{tabular}} &
  \textbf{\begin{tabular}[c]{@{}c@{}}FFT\\ loss\end{tabular}} &
  \textbf{PSNR} &
  \textbf{\begin{tabular}[c]{@{}c@{}}SSIM\\ ($\times$100)\end{tabular}} \\ \hline
\checkmark &
   &
   &
   &
   &
   &
   &
   &
  \checkmark &
  \checkmark &
  \checkmark &
  42.32 &
  96.11 \\ \cline{12-13} 
 &
  \checkmark &
   &
   &
   &
   &
   &
   &
  \checkmark &
  \checkmark &
  \checkmark &
  40.93 &
  95.56 \\ \cline{12-13} 
\checkmark &
  \checkmark &
   &
   &
   &
   &
   &
   &
  \checkmark &
  \checkmark &
  \checkmark &
  42.50 &
  97.12 \\ \cline{12-13} 
\checkmark &
  \checkmark &
  \checkmark &
   &
   &
   &
   &
   &
  \checkmark &
  \checkmark &
  \checkmark &
  42.70 &
  97.24 \\ \cline{12-13} 
\checkmark &
  \checkmark &
  \checkmark &
  \checkmark &
   &
   &
   &
   &
  \checkmark &
  \checkmark &
  \checkmark &
  42.80 &
  97.19 \\ \cline{12-13} 
\rowcolor[HTML]{EFEFEF} 
\checkmark &
  \checkmark &
  \checkmark &
  \checkmark &
  \checkmark &
   &
   &
   &
  \checkmark &
  \checkmark &
  \checkmark &
  \cellcolor[HTML]{C1FED2}\textbf{43.58} &
  \cellcolor[HTML]{C1FED2}\textbf{97.67} \\ \cline{12-13} 
 &
  \checkmark &
  \checkmark &
  \checkmark &
  \checkmark &
  \checkmark &
   &
   &
   &
   &
   &
  43.56 &
  97.64 \\ \cline{12-13} 
\checkmark &
   &
  \checkmark &
  \checkmark &
  \checkmark &
   &
  \checkmark &
   &
   &
   &
   &
  43.49 &
  97.66 \\ \cline{12-13} 
\checkmark &
   &
  \checkmark &
  \checkmark &
  \checkmark &
   &
   &
  \checkmark &
   &
   &
   &
  43.42 &
  97.63 \\ \cline{12-13} 
\checkmark &
  \checkmark &
  \checkmark &
  \checkmark &
  \checkmark &
   &
   &
   &
  \checkmark &
   &
  \checkmark &
  43.06 &
  97.35 \\ \cline{12-13} 
\checkmark &
  \checkmark &
  \checkmark &
  \checkmark &
  \checkmark &
   &
   &
   &
  \checkmark &
  \checkmark &
   &
  43.34 &
  97.59 \\ \cline{12-13} 
\checkmark &
  \checkmark &
  \checkmark &
  \checkmark &
  \checkmark &
   &
   &
   &
   &
  \checkmark &
  \checkmark &
  41.72 &
  97.62 \\ \hline
\end{tabular}
}
\end{table}

%% file: tables/jpeg2000.tex
\begin{table}[t]
\centering
\caption{Quantitative results using JPEG 2000 \cite{marcellin2000overview} for image compression instead of JPEG. We report average PSNR, SSIM, and MS-SSIM \cite{ssim, ms_ssim} between uncompressed raw images and the decoded outputs. Compression ratio (CR) is measured relative to the 16-bit RGB PNG, and BPP reflects the compressed file size. Best results are highlighted in \textcolor{green}{\textbf{green}}.\label{tab:jpeg-2000}}
\vspace{-1mm}
\scalebox{0.7}{
\begin{tabular}{|l|c|c|c|c|c|}
\hline
\multicolumn{1}{|c|}{\textbf{Pre-/post-processing}}
& \multicolumn{1}{c|}{\textbf{PSNR}} &
  \multicolumn{1}{c|}{\textbf{\begin{tabular}[c]{@{}c@{}}SSIM\\ ($\times$100)\end{tabular}}} &
  \multicolumn{1}{c|}{\textbf{\begin{tabular}[c]{@{}c@{}}MS-SSIM\\ ($\times$100)\end{tabular}}} &
  \multicolumn{1}{c|}{\textbf{BPP}} &
  \textbf{CR} \\
\hline
None & \multicolumn{1}{c|}{52.96} & \multicolumn{1}{c|}{99.66} & \multicolumn{1}{c|}{99.96} & \multicolumn{1}{c|}{7.58} & 3.76
\\ \hline

Ours (trained on quality 50) & \multicolumn{1}{c|}{56.96} & \multicolumn{1}{c|}{99.91} & \multicolumn{1}{c|}{\cellcolor{best}\textbf{99.99}} & \multicolumn{1}{c|}{11.13} & 2.51 \\ \hdashline

Ours (trained on quality 75) & \multicolumn{1}{c|}{\cellcolor{best}\textbf{57.64}} & \multicolumn{1}{c|}{\cellcolor{best}\textbf{99.94}} & \multicolumn{1}{c|}{\cellcolor{best}\textbf{99.99}} & \multicolumn{1}{c|}{10.84} & 2.58 \\ \hline

\end{tabular}
}
\end{table}

%% file: tables/results_supp_2.tex
\begin{table}[t]
\centering
\caption{Quantitative results on cross-camera generalization to unseen cameras during training. We report average PSNR and SSIM \cite{ssim, ms_ssim} between the uncompressed raw images and decoded output at JPEG quality 100. BPP denotes JPEG file size, and CR denotes the compression ratio with respect to the uncompressed 16-bit RGB PNG. Best results are highlighted in \textcolor{green}{\textbf{green}}.\label{tab:cross-camera-quality-100-supp}}\vspace{-1mm}
\scalebox{0.7}{
\begin{tabular}{|c|c|c|c|c|}
\hline
\multicolumn{5}{|c|}{\cellcolor[HTML]{e6e5e1}\textbf{S7 dataset \cite{S7} (220 images from 1 smartphone camera)}} \\
\multirow{2}{*}{\textbf{Method}} &
  \multicolumn{4}{c|}{\cellcolor[HTML]{fff0dd}\textbf{Quality = 100}} \\
\cline{2-5} & \textbf{PSNR} &   \multicolumn{1}{c|}{\textbf{\begin{tabular}[c]{@{}c@{}}SSIM\\ ($\times$100)\end{tabular}}} & \textbf{BPP} & \textbf{CR} \\
\hline
JPEG
& 
\multicolumn{1}{c|}{47.84} & \multicolumn{1}{c|}{98.58} & \multicolumn{1}{c|}{4.57} & 8.89
\\ \hline
JPEG + raw$\leftrightarrow$sRGB
& \multicolumn{1}{c|}{45.13} & \multicolumn{1}{c|}{98.92} & \multicolumn{1}{c|}{7.70} & 5.23  
\\ \hline
JPEG + fixed gamma
& \multicolumn{1}{c|}{50.95} & \multicolumn{1}{c|}{99.31} & \multicolumn{1}{c|}{6.48} & 6.29
\\ \hdashline
 Raw-JPEG Adapter (w/ $\text{ }$ DCT)  
 & \multicolumn{1}{c|}{48.62} & \multicolumn{1}{c|}{\cellcolor{best}\textbf{99.32}} & \multicolumn{1}{c|}{6.36} & 6.37
\\ \hline
Raw-JPEG Adapter (w/o DCT)   
& 
\multicolumn{1}{c|}{\cellcolor{best}\textbf{50.96}} & \multicolumn{1}{c|}{99.31} & \multicolumn{1}{c|}{6.57} & 6.21
\\ \hline
\multicolumn{5}{|c|}{\cellcolor[HTML]{e6e5e1}\textbf{MIT-Adobe 5K dataset \cite{Adobe5K} (5,000 images from 35 DSLR cameras)}} \\
\hline
JPEG
& \multicolumn{1}{c|}{48.35} & \multicolumn{1}{c|}{98.60} & \multicolumn{1}{c|}{3.36} & 11.58
\\ \hline
JPEG + raw$\leftrightarrow$sRGB
& \multicolumn{1}{c|}{43.26} & \multicolumn{1}{c|}{98.70} & \multicolumn{1}{c|}{5.47} & 6.84  
\\ \hline
JPEG + fixed gamma
& \multicolumn{1}{c|}{51.99} & \multicolumn{1}{c|}{\cellcolor{best}\textbf{99.60}} & \multicolumn{1}{c|}{4.71} & 7.95
\\ \hdashline
 Raw-JPEG Adapter (w/ $\text{ }$ DCT)  
 & \multicolumn{1}{c|}{48.17} & \multicolumn{1}{c|}{99.59} & \multicolumn{1}{c|}{4.68} & 7.99
\\ \hline
Raw-JPEG Adapter (w/o DCT)   
& \multicolumn{1}{c|}{\cellcolor{best}\textbf{52.00}} & \multicolumn{1}{c|}{\cellcolor{best}\textbf{99.60}} & \multicolumn{1}{c|}{4.76} & 7.87 
\\ \hline
\multicolumn{5}{|c|}{\cellcolor[HTML]{e6e5e1}\textbf{NUS dataset \cite{NUS} (1,736 images from 8 DSLR cameras)}} \\
\hline
JPEG
& \multicolumn{1}{c|}{45.74} & \multicolumn{1}{c|}{97.82} & \multicolumn{1}{c|}{3.58} & 9.55
\\ \hline
JPEG + raw$\leftrightarrow$sRGB
& \multicolumn{1}{c|}{45.52} & \multicolumn{1}{c|}{99.18} & \multicolumn{1}{c|}{5.10} & 5.10 
\\ \hline
JPEG + fixed gamma
& \multicolumn{1}{c|}{49.25} & \multicolumn{1}{c|}{99.21} & \multicolumn{1}{c|}{4.84} & 6.90
\\ \hdashline
 Raw-JPEG Adapter (w/ $\text{ }$ DCT)  
 & \multicolumn{1}{c|}{47.62} & \multicolumn{1}{c|}{\cellcolor{best}\textbf{99.23}} & \multicolumn{1}{c|}{4.83} & 6.92
\\ \hline
Raw-JPEG Adapter (w/o DCT)   
& \multicolumn{1}{c|}{\cellcolor{best}\textbf{49.36}} & \multicolumn{1}{c|}{99.22} & \multicolumn{1}{c|}{4.95} & 6.76
\\ \hline
\end{tabular}
}
\end{table}

%% file: tables/results_srgb.tex
\begin{table}[t]
\centering
\caption{Quantitative results on sRGB images from the Kodak set \cite{Kodak1993}. We report the average PSNR, SSIM, and MS-SSIM \cite{ssim, ms_ssim} between the uncompressed sRGB images and the decoded outputs across different JPEG quality levels. For our method, we provide results from the model trained on sRGB images, both with and without the DCT component. Best results are highlighted in \textcolor{green}{\textbf{green}}. \label{tab:srgb}}\vspace{-2mm}
\scalebox{0.65}{
\begin{tabular}{|c|ccc|ccc|ccc|}
\hline
\multirow{2}{*}{\textbf{Method}} &
  \multicolumn{3}{c|}{\cellcolor[HTML]{fed18d}\textbf{Quality = 25}} &
  \multicolumn{3}{c|}{\cellcolor[HTML]{ff9484}\textbf{Quality = 50}} &
  \multicolumn{3}{c|}{\cellcolor[HTML]{d5effe}\textbf{Quality = 75}} \\
\cline{2-10}
& \multicolumn{1}{c|}{\textbf{PSNR}} &
  \multicolumn{1}{c|}{\textbf{\begin{tabular}[c]{@{}c@{}}SSIM\\ ($\times$100)\end{tabular}}} &
  \multicolumn{1}{c|}{\textbf{\begin{tabular}[c]{@{}c@{}}MS-SSIM\\ ($\times$100)\end{tabular}}} &
  \multicolumn{1}{c|}{\textbf{PSNR}} &
  \multicolumn{1}{c|}{\textbf{\begin{tabular}[c]{@{}c@{}}SSIM\\ ($\times$100)\end{tabular}}} &
  \multicolumn{1}{c|}{\textbf{\begin{tabular}[c]{@{}c@{}}MS-SSIM\\ ($\times$100)\end{tabular}}} &
  \multicolumn{1}{c|}{\textbf{PSNR}} &
  \multicolumn{1}{c|}{\textbf{\begin{tabular}[c]{@{}c@{}}SSIM\\ ($\times$100)\end{tabular}}} &
  \textbf{\begin{tabular}[c]{@{}c@{}}MS-SSIM\\ ($\times$100)\end{tabular}} \\
\hline
\textbf{JPEG} &
\multicolumn{1}{c|}{29.89} & \multicolumn{1}{c|}{84.97} & \multicolumn{1}{c|}{95.61} & \multicolumn{1}{c|}{32.17} & \multicolumn{1}{c|}{89.97} & \multicolumn{1}{c|}{97.69} & \multicolumn{1}{c|}{34.52} & \multicolumn{1}{c|}{93.32} & 98.66  \\
\hline

\textbf{JPEG + fixed gamma} &
\multicolumn{1}{c|}{28.81} & \multicolumn{1}{c|}{82.80} & \multicolumn{1}{c|}{94.48} & \multicolumn{1}{c|}{31.03} & \multicolumn{1}{c|}{88.34} & \multicolumn{1}{c|}{97.10} & \multicolumn{1}{c|}{33.18} & \multicolumn{1}{c|}{92.07} & 98.32 
  \\ \hdashline

\textbf{JPEG + ours (w/o DCT)} &
\multicolumn{1}{c|}{30.25} & \multicolumn{1}{c|}{87.89} & \multicolumn{1}{c|}{96.85} & \multicolumn{1}{c|}{31.21} & \multicolumn{1}{c|}{90.35} & \multicolumn{1}{c|}{97.64} & \multicolumn{1}{c|}{\cellcolor{best}\textbf{34.83}} & \multicolumn{1}{c|}{94.33} & 98.92  \\
\hline

\textbf{JPEG + ours (w/$\text{ }$ DCT)} &
\multicolumn{1}{c|}{\cellcolor{best}\textbf{30.31}} & \multicolumn{1}{c|}{\cellcolor{best}\textbf{87.98}} & \multicolumn{1}{c|}{\cellcolor{best}\textbf{96.90}} & \multicolumn{1}{c|}{\cellcolor{best}\textbf{32.39}} & \multicolumn{1}{c|}{\cellcolor{best}\textbf{91.61}} & \multicolumn{1}{c|}{\cellcolor{best}\textbf{98.19}} & \multicolumn{1}{c|}{34.75} & \multicolumn{1}{c|}{\cellcolor{best}\textbf{94.40}} & \cellcolor{best}\textbf{98.93}  \\
\hline

\end{tabular}

}
\end{table}

%% file: tables/neural_isp_tab_1.tex
\begin{table*}[!h]
\centering
\caption{Quantitative results on raw-to-sRGB rendering using the S24 test set \cite{S24}. We report average PSNR and SSIM \cite{ssim, ms_ssim} between the ground-truth sRGB (picture style \#0) and outputs from LiteISP \cite{lite-isp} and Invertible ISP \cite{xing21invertible}, both trained on the S24 training set. Each row corresponds to a different input: (1) uncompressed raw (27.78 BPP), (2) JPEG-compressed raw (0.25/0.36/0.56/1.67 BPP for qualities 25/50/75/95), (3) JPEG raw with fixed gamma (0.30/0.47/0.74/2.16 BPP), and (4) JPEG raw processed with our Raw-JPEG Adapter (0.47/0.69/1.01/2.46 BPP). Results for our method are shown with and without the DCT component. Best scores are highlighted in \textcolor{green}{\textbf{green}}. \label{tab:neural-isp-1}}
\scalebox{0.65}{
\begin{tabular}{|lclclclll|}

\hline
\multicolumn{9}{|c|}{\cellcolor[HTML]{E6E5E1}LiteISP \cite{lite-isp}} \\ \hline

\multicolumn{1}{|l|}{} &
  \multicolumn{2}{c|}{\cellcolor[HTML]{fed18d}\textbf{Quality = 25}} &
  \multicolumn{2}{c|}{\cellcolor[HTML]{ff9484}\textbf{Quality = 50}} &
  \multicolumn{2}{c|}{\cellcolor[HTML]{d5effe}\textbf{Quality = 75}} &
  \multicolumn{2}{c|}{\cellcolor[HTML]{CBCEFB}\textbf{Quality = 95}} \\ \cline{2-9} 
\multicolumn{1}{|l|}{\multirow{-2}{*}{\textbf{Input raw}}} &
  \multicolumn{1}{c|}{\textbf{PSNR}} &
  \multicolumn{1}{c|}{\textbf{\begin{tabular}[c]{@{}c@{}}SSIM \\ ($\times$100)\end{tabular}}} &
  \multicolumn{1}{c|}{\textbf{PSNR}} &
  \multicolumn{1}{c|}{\textbf{\begin{tabular}[c]{@{}c@{}}SSIM \\ ($\times$100)\end{tabular}}} &
  \multicolumn{1}{c|}{\textbf{PSNR}} &
  \multicolumn{1}{c|}{\textbf{\begin{tabular}[c]{@{}c@{}}SSIM \\ ($\times$100)\end{tabular}}} &
  \multicolumn{1}{c|}{\textbf{PSNR}} &
  \multicolumn{1}{c|}{\textbf{\begin{tabular}[c]{@{}c@{}}SSIM \\ ($\times$100)\end{tabular}}} \\ \hline
\multicolumn{1}{|l|}{Uncompressed} &
\multicolumn{1}{c|}{25.95} & \multicolumn{1}{c|}{90.41} & \multicolumn{1}{c|}{25.95} & \multicolumn{1}{c|}{90.41} & \multicolumn{1}{c|}{25.95} & \multicolumn{1}{c|}{90.41} & \multicolumn{1}{c|}{25.95} & 90.41 
   \\ \hline
\multicolumn{1}{|l|}{JPEG} &
\multicolumn{1}{c|}{17.48} & \multicolumn{1}{c|}{62.69} & \multicolumn{1}{c|}{19.40} & \multicolumn{1}{c|}{72.64} & \multicolumn{1}{c|}{20.78} & \multicolumn{1}{c|}{78.99} & \multicolumn{1}{c|}{23.69} & 85.94 
   \\ \hline
\multicolumn{1}{|l|}{JPEG + fixed gamma} &
\multicolumn{1}{c|}{21.50} & \multicolumn{1}{c|}{73.39} & \multicolumn{1}{c|}{22.94} & \multicolumn{1}{c|}{81.39} & \multicolumn{1}{c|}{24.00} & \multicolumn{1}{c|}{85.80} & \multicolumn{1}{c|}{25.38} & 90.04 
 \\ \hdashline
\multicolumn{1}{|l|}{Raw-JPEG Adapter (w/o DCT)} &
\multicolumn{1}{c|}{\cellcolor{best}\textbf{22.77}} & \multicolumn{1}{c|}{80.32} & \multicolumn{1}{c|}{\cellcolor{best}\textbf{23.83}} & \multicolumn{1}{c|}{\cellcolor{best}\textbf{85.05}} & \multicolumn{1}{c|}{\cellcolor{best}\textbf{24.60}} & \multicolumn{1}{c|}{87.99} & \multicolumn{1}{c|}{\cellcolor{best}\textbf{25.53}} & \cellcolor{best}\textbf{90.26} 
   \\ \hline
\multicolumn{1}{|l|}{Raw-JPEG Adapter (w/$\text{ }$ DCT)} &
\multicolumn{1}{c|}{22.53} & \multicolumn{1}{c|}{\cellcolor{best}\textbf{80.43}} & \multicolumn{1}{c|}{23.32} & \multicolumn{1}{c|}{84.78} & \multicolumn{1}{c|}{24.50} & \multicolumn{1}{c|}{\cellcolor{best}\textbf{88.11}} & \multicolumn{1}{c|}{25.24} & 90.03 
   \\ \hline

\multicolumn{9}{|c|}{\cellcolor[HTML]{E6E5E1}Invertible ISP \cite{xing21invertible}} \\ \hline

\multicolumn{1}{|l|}{Uncompressed} &
\multicolumn{1}{c|}{22.87} & \multicolumn{1}{c|}{81.97} & \multicolumn{1}{c|}{22.87} & \multicolumn{1}{c|}{81.97} & \multicolumn{1}{c|}{22.87} & \multicolumn{1}{c|}{81.97} & \multicolumn{1}{c|}{22.87} & 81.97 
   \\ \hline
\multicolumn{1}{|l|}{JPEG} &
\multicolumn{1}{c|}{18.91} & \multicolumn{1}{c|}{58.51} & \multicolumn{1}{c|}{20.56} & \multicolumn{1}{c|}{66.66} & \multicolumn{1}{c|}{21.42} & \multicolumn{1}{c|}{71.90} & \multicolumn{1}{c|}{22.20} & 74.30 
   \\ \hline
\multicolumn{1}{|l|}{JPEG + fixed gamma} &
\multicolumn{1}{c|}{21.18} & \multicolumn{1}{c|}{71.31} & \multicolumn{1}{c|}{22.01} & \multicolumn{1}{c|}{77.41} & \multicolumn{1}{c|}{22.42} & \multicolumn{1}{c|}{80.17} & \multicolumn{1}{c|}{22.81} & 80.61 
 \\ \hdashline
\multicolumn{1}{|l|}{Raw-JPEG Adapter (w/o DCT)} &
\multicolumn{1}{c|}{\cellcolor{best}\textbf{21.71}} & \multicolumn{1}{c|}{76.56} & \multicolumn{1}{c|}{\cellcolor{best}\textbf{22.26}} & \multicolumn{1}{c|}{\cellcolor{best}\textbf{79.94}} & \multicolumn{1}{c|}{\cellcolor{best}\textbf{22.54}} & \multicolumn{1}{c|}{81.19} & \multicolumn{1}{c|}{\cellcolor{best}\textbf{22.82}} & \cellcolor{best}\textbf{80.91}
   \\ \hline
\multicolumn{1}{|l|}{Raw-JPEG Adapter (w/$\text{ }$ DCT)} &
\multicolumn{1}{c|}{21.43} & \multicolumn{1}{c|}{\cellcolor{best}\textbf{76.81}} & \multicolumn{1}{c|}{21.87} & \multicolumn{1}{c|}{79.79} & \multicolumn{1}{c|}{22.44} & \multicolumn{1}{c|}{\cellcolor{best}\textbf{81.20}} & \multicolumn{1}{c|}{22.56} & 80.69 
   \\ \hline

\end{tabular}
}
\end{table*}

%% file: tables/neural_isp_tab_2.tex
\begin{table*}[!h]
\centering
\caption{Quantitative results on image re-rendering using the S24 test set \cite{S24}. We report average PSNR and SSIM \cite{ssim, ms_ssim} between ground-truth sRGB images with different picture styles and the outputs from Invertible ISP models \cite{xing21invertible}, each trained to reproduce a specific target style. Each row corresponds to a different input: (1) a reconstructed raw image obtained by inverting a previously rendered sRGB image using an Invertible ISP trained on either Style \#0 (first group in the table) or Style \#5 (second group), and (2) a raw image saved using our Raw-JPEG Adapter, which achieves a significantly smaller file size compared to storing in DNG or PNG-16 format. Unlike the Invertible ISP, which requires $\sim$7 seconds on GPU to reconstruct a raw image, our method provides an affordable raw file size without requiring reconstruction. Best scores are highlighted in \textcolor{green}{\textbf{green}}.\label{tab:neural-isp-2}}
\scalebox{0.7}{
\begin{tabular}{|ccccccccccc|}
\hline
\multicolumn{1}{|c|}{\multirow{2}{*}{\textbf{Input raw image}}} &
  \multicolumn{10}{c|}{\cellcolor[HTML]{ebe7e7}\textbf{Reconstructed raw from sRGB (Style \#0)}} \\ \cline{2-11} 
\multicolumn{1}{|c|}{} &
  \multicolumn{2}{c|}{\cellcolor[HTML]{fed18d}\textbf{Style \#1}} &
  \multicolumn{2}{c|}{\cellcolor[HTML]{ff9484}\textbf{Style \#2}} &
  \multicolumn{2}{c|}{\cellcolor[HTML]{d5effe}\textbf{Style \#3}} &
  \multicolumn{2}{c|}{\cellcolor[HTML]{CBCEFB}\textbf{Style \#4}} &
  \multicolumn{2}{c|}{\cellcolor[HTML]{fff0d9}\textbf{Style \#5}} \\ \cline{2-11} 
\multicolumn{1}{|c|}{} &
  \multicolumn{1}{c|}{\textbf{PSNR}} &
  \multicolumn{1}{c|}{\textbf{SSIM}} &
  \multicolumn{1}{c|}{\textbf{PSNR}} &
  \multicolumn{1}{c|}{\textbf{SSIM}} &
  \multicolumn{1}{c|}{\textbf{PSNR}} &
  \multicolumn{1}{c|}{\textbf{SSIM}} &
  \multicolumn{1}{c|}{\textbf{PSNR}} &
  \multicolumn{1}{c|}{\textbf{SSIM}} &
  \multicolumn{1}{c|}{\textbf{PSNR}} &
  \textbf{SSIM} \\ \hline
\multicolumn{1}{|c|}{Reconstructed via Invertible ISP \cite{xing21invertible}} &
\multicolumn{1}{c|}{21.38} & \multicolumn{1}{c|}{80.99} & \multicolumn{1}{c|}{26.17} & \multicolumn{1}{c|}{84.26} & \multicolumn{1}{c|}{\cellcolor{best}\textbf{23.82}} & \multicolumn{1}{c|}{\cellcolor{best}\textbf{85.05}} & \multicolumn{1}{c|}{\cellcolor{best}\textbf{23.30}} & \multicolumn{1}{c|}{\cellcolor{best}\textbf{84.21}} & \multicolumn{1}{c|}{24.89} & 86.96 \\\hdashline
\multicolumn{1}{|c|}{Raw-JPEG Adapter} &
\multicolumn{1}{c|}{\cellcolor{best}\textbf{23.47}} & \multicolumn{1}{c|}{\cellcolor{best}\textbf{83.09}} & \multicolumn{1}{c|}{\cellcolor{best}\textbf{26.27}} & \multicolumn{1}{c|}{\cellcolor{best}\textbf{84.83}} & \multicolumn{1}{c|}{23.80} & \multicolumn{1}{c|}{84.85} & \multicolumn{1}{c|}{23.28} & \multicolumn{1}{c|}{83.61} & \multicolumn{1}{c|}{\cellcolor{best}\textbf{24.90}} & \cellcolor{best}\textbf{87.00} \\ \hline
\multirow{2}{*}{} &
  \multicolumn{10}{|c|}{\cellcolor[HTML]{ebe7e7}\textbf{Reconstructed raw from sRGB (Style \#5)}} \\ \cline{2-11} 
 &
  \multicolumn{2}{|c|}{\cellcolor[HTML]{fed18d}\textbf{Style \#0}} &
  \multicolumn{2}{c|}{\cellcolor[HTML]{ff9484}\textbf{Style \#1}} &
  \multicolumn{2}{c|}{\cellcolor[HTML]{d5effe}\textbf{Style \#2}} &
  \multicolumn{2}{c|}{\cellcolor[HTML]{CBCEFB}\textbf{Style \#3}} &
  \multicolumn{2}{c|}{\cellcolor[HTML]{fff0d9}\textbf{Style \#4}} \\ \cline{2-11} 
 &
  \multicolumn{1}{|c|}{\textbf{PSNR}} &
  \multicolumn{1}{c|}{\textbf{SSIM}} &
  \multicolumn{1}{c|}{\textbf{PSNR}} &
  \multicolumn{1}{c|}{\textbf{SSIM}} &
  \multicolumn{1}{c|}{\textbf{PSNR}} &
  \multicolumn{1}{c|}{\textbf{SSIM}} &
  \multicolumn{1}{c|}{\textbf{PSNR}} &
  \multicolumn{1}{c|}{\textbf{SSIM}} &
  \multicolumn{1}{c|}{\textbf{PSNR}} &
  \textbf{SSIM} \\ \hline
\multicolumn{1}{|c|}{Reconstructed via Invertible ISP \cite{xing21invertible}} &
\multicolumn{1}{c|}{\cellcolor{best}\textbf{22.82}} & \multicolumn{1}{c|}{80.12} & \multicolumn{1}{c|}{22.85} & \multicolumn{1}{c|}{82.78} & \multicolumn{1}{c|}{25.91} & \multicolumn{1}{c|}{84.12} & \multicolumn{1}{c|}{23.39} & \multicolumn{1}{c|}{84.03} & \multicolumn{1}{c|}{22.85} & 82.78  \\ \hdashline
\multicolumn{1}{|c|}{Raw-JPEG Adapter} &
 \multicolumn{1}{c|}{\cellcolor{best}\textbf{22.82}} & \multicolumn{1}{c|}{\cellcolor{best}\textbf{80.91}} & \multicolumn{1}{c|}{\cellcolor{best}\textbf{23.47}} & \multicolumn{1}{c|}{\cellcolor{best}\textbf{83.09}} & \multicolumn{1}{c|}{\cellcolor{best}\textbf{26.27}} & \multicolumn{1}{c|}{\cellcolor{best}\textbf{84.83}} & \multicolumn{1}{c|}{\cellcolor{best}\textbf{23.80}} & \multicolumn{1}{c|}{\cellcolor{best}\textbf{84.85}} & \multicolumn{1}{c|}{\cellcolor{best}\textbf{23.28}} & \cellcolor{best}\textbf{83.61}  \\ \hline
 
\end{tabular}}
\end{table*}